\documentclass{article}
\usepackage{arxiv}

\usepackage{graphicx}
\usepackage{amsmath,amssymb} % define this before the line numbering.
\usepackage{xcolor}
\usepackage{soul}
\usepackage{xspace}
\usepackage{epsfig}
\usepackage{times}
\usepackage{blindtext}
\usepackage{mathtools}
\usepackage[titlenumbered,ruled,linesnumbered]{algorithm2e}

\makeatletter
\DeclareRobustCommand\onedot{\futurelet\@let@token\@onedot}
\def\@onedot{\ifx\@let@token.\else.\null\fi\xspace}

\def\etal{\emph{et al}\onedot}
\makeatother

\def\w{3.1cm}
\def\v{3.7cm}

\SetKwComment{Comment}{$\triangleright$\ }{}
\newcommand{\argmin}{\operatornamewithlimits{argmin}}
\graphicspath{{./figures/}}

\title{Super-Trajectories: A Compact Yet Rich Video Representation}

\author{
  Ijaz~Akhter \\
  National University of Singapore\\
  \texttt{eleijaz@nus.edu.sg} \\
   \And
  Cheong~Loong~Fah\\
  National University of Singapore\\
  \texttt{eleclf@nus.edu.sg} \\   
  \And
Richard~Hartley \\
  Australian National University\\
  \texttt{Richard.Hartley@anu.edu.au} \\
}

\begin{document}
\maketitle

\begin{abstract}
We propose a new video representation in terms of an over-segmentation of dense trajectories covering the whole video. Trajectories are often used to encode long-temporal information in several computer vision applications. Similar to temporal superpixels, a temporal slice of super-trajectories are superpixels, but the later contains more information because it maintains the long dense pixel-wise tracking information as well. The main challenge in using trajectories for any application, is the accumulation of tracking error in the trajectory construction. For our problem, this results in disconnected superpixels. We exploit constraints for edges in addition to trajectory based color and position similarity. Analogous to superpixels as a preprocessing tool for images, the proposed representation has its applications for videos, especially in trajectory based video analysis.
\end{abstract}

% keywords can be removed

\section{Introduction}
\begin{figure}
\centering
\includegraphics[width=7cm]{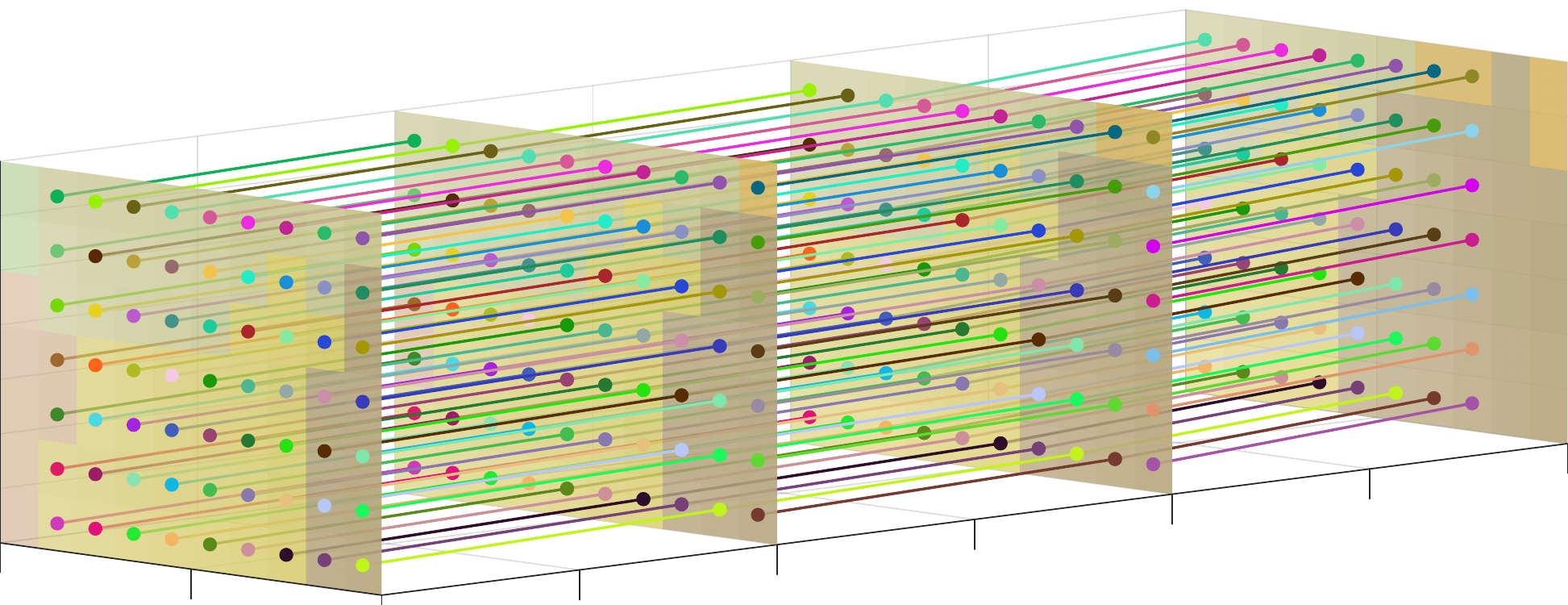}
\hspace{2mm}
\includegraphics[width=7cm]{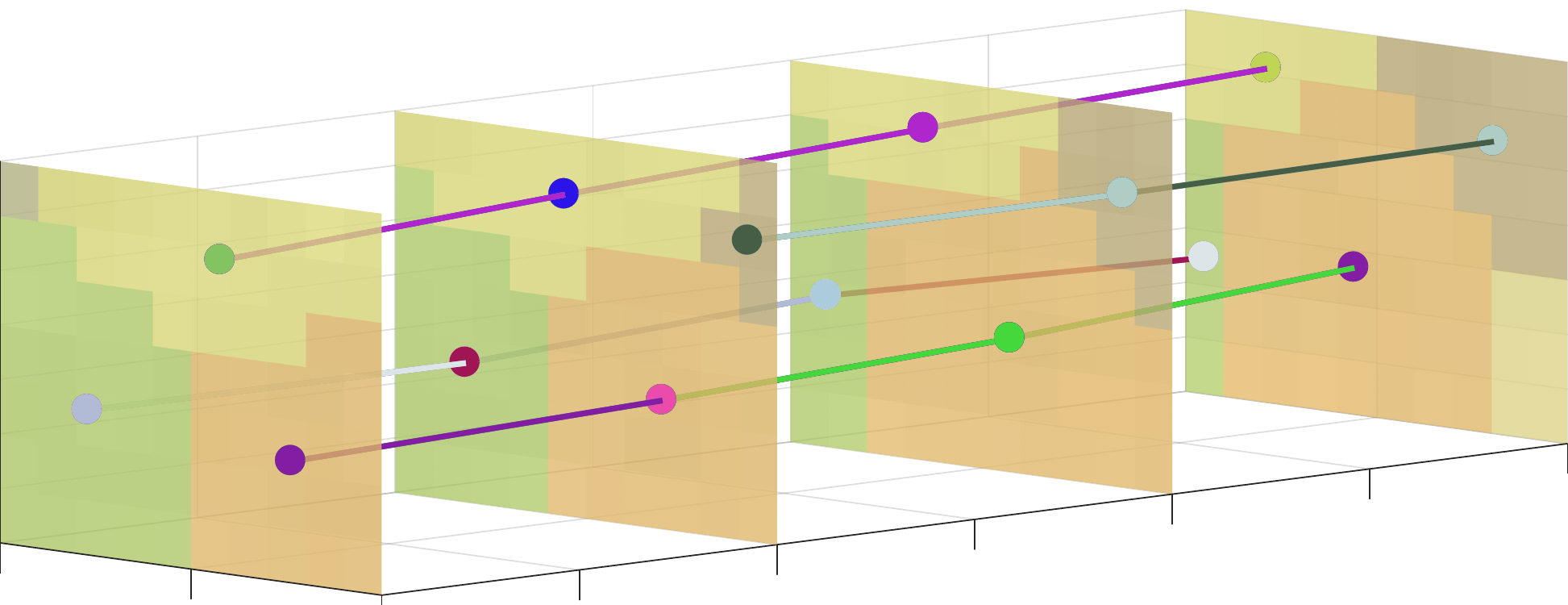}
\caption{ An illustration of the difference between super-trajectories (left) and temporal superpixels (right). Super-trajectories segmentation is a clustering of dense trajectories (shown with the color lines), where each trajectory bears a single cluster label (shown with the grid colors). Whereas, temporal superpixels assign labels at pixel-level and other than the center, they do not provide the coordinates of any other trajectory. Both of these representations provide tracking of superpixels.}\label{fig:teaser}
\end{figure}

Trajectories --- 2D tracks of feature points along time, are often used in computer vision to model long temporal information in a video. A majority of these methods involve sparse trajectories\cite{yan2006general,ochs2011object,fragkiadaki2012video}, though dense trajectories has also been explored \cite{wang2013dense,wang2017super}. Many inference problems require finding pair-wise similarities or affinities between trajectories \cite{rao2008motion,liu2010robust}. However, finding pair-wise affinites among a large neighbourhood in dense trajectories is often not tractable. Temporal superpixels can be used to find a compact representation of the video but they do not keep the trajectory information and by using only the central trajectory, long pixel-level tracking information is lost. In this paper, we propose super-trajectories as an over-segmentation of a video into clusters of dense trajectories. The goal is to have them connected as temporal superpixels. But in contrast to temporal superpixels, where labels are assigned at pixel level, in our approach, each trajectory bears a single label (See Fig \ref{fig:teaser}). With our approach, the affinity between all pairs of trajectories in two super-trajectories can be found. More importantly, the segmentation into clusters allows us to perform intelligent sampling and save computation time. This is especially crucial for those motion segmentation approaches based on the hypothesis-and-test paradigm, where one needs to hypothesize homographies or fundamental matrices based on a minimal set of matches between all pairs of frames. Leveraging the cluster information will significantly raise the chance of getting all inliers for the motion hypothesis. Once a robust set of trajectories have been used to form accurate homographies or fundamental matrices, the latter can be fit on those less accurate trajectories.

Finding accurate dense trajectories covering every pixel in the video is a challenging problem. To accommodate new and occluding scene and avoid accumulation of tracking error, we propose a consistency test for optical flow by exploiting the color and the edge boundaries in the image sequence. We find trajectories by the composition of optical flow based warps and use the proposed consistency test to decide whether to break or continue a trajectory. This usually results in a large number of trajectories (up to millions) of both long and short durations (due to occlusion). Clustering these trajectories is a classification problem of big and highly incomplete data of high dimensions and is quite challenging. Please note that the spectral clustering based clustering methods \cite{ng2002spectral} mostly require estimation of an affinity matrix for every pair of trajectories and are not tractable for dense trajectories. Second, due to the accumulation of tracking errors, even the tracks corresponding to two nearby points that should be grouped together, can have large distances in some of the images. Consequently, many neighbouring trajectories would only be partially connected and ensuring that super-trajectories would result in connected superpixels throughout the video is very hard. This requires us to redefine the notion of neighbours and connectivity for trajectories. The goal of the proposed method is to minimize disconnections of the super-trajectories, or effectively those of the underlying superpixels.

Apart from the proposed trajectory estimation, the main contribution of the paper is an iterative algorithm, where each iteration reduces the number of disconnections until convergence. Part of each iteration is an adaptation of the recently proposed, non-iterative superpixeling algorithm SNIC \cite{achanta2017snic}, for trajectories. Specfically, we define for trajectories color and position based similarity; we also propose edge based similarity constraints for trajectories for better localization of superpixels. In contrast to superpixels based methods, the final post-processing step to filter out small isolated regions of labels is more complicated for trajectories than pixels because neighbouring trajectories are often only partially connected. The proposed post-processing step relabels the isolated trajectories to minimize the disconnections. The proposed method is able to track superpixels much longer than the previous methods, while also slightly improving on the under-segmentation error on Chen dataset \cite{chen2010propagating}.

%In contrast to superpixels based methods, the initialization for super-trajectories is nontrivial. We initialize cluster centers with evenly separated trajectories such that the remaining trajectories lie within the average diameter of the superpixels, as given by the user.  Similarly the final post-processing step to filter out small isolated regions of labels is more complicated for trajectories than pixels because neighbouring trajectories are often only partially connected. The proposed post-processing step relabels the isolated trajectories to minimize the disconnections. 

The  design choice of including the whole trajectory in a super-trajectory comes at a cost. Due to accumulation of tracking error and drift, the boundaries of the corresponding superpixels, sometimes, cannot be accurately localized. As a result, super-trajectories segmentation accuracy and boundary recall are slightly worse than the existing temporal superpixels methods \cite{chang2013video,lee2017contour}. Nevertheless, this problem happens only for a very small number of super-trajectories; the rest of them are perceptually of similar quality to the temporal superpixels while carrying more information in the form of dense pixel-wise tracking. With the help of more accurate optical flow, super-trajectories segmentation should further improve in accuracy.

%=====================================================================================
\section{Related Work}
Image segmentation into superpixels is a widely studied problem in Computer Vision. Here we only discuss some of prominent works in this area. Normalized cuts algorithm, by Shi and Malik, uses contour and texture cues to recursively partition the image using a pixel graph \cite{shi2000normalized}. Meanshift, proposed by Comaniciu and Meer, is a local mode seeking algorithm on the color and position space to find segment of the image \cite{comaniciu2002mean}. Quickshift, by Vedaldi and Soatto, is also a mode seeking scheme but more efficient than meanshift \cite{vedaldi2008quick}. SEEDS, by Van den Bergh \etal, is a coarse-to-fine method to refine superpixel boundaries through an energy-driven sampling \cite{van2012seeds}. SLIC, by Achanta \etal, is an optimized K-means clustering algorithm on color and position features \cite{achanta2012slic}. A more comprehensive list of superpixel clustering algorithms and their evaluation is available in \cite{stutz2017superpixels}.

In contrast to single-frame superpixels, temporal superpixels are not extensively studied. Extending superpixels along time requires enforcing temporal continuity. Several temporal superpixels method require optical flow but dealing with inaccuracies in optical flow is not a trivial problem. Van den Bergh \etal proposed an extension of SEEDS to get temporal superpixel segmentation of video in an online fashion \cite{van2013online}. Reso \etal used K-means algorithm in a temporal sliding window fashion to impose temporal consistency \cite{reso2013temporally}. Chang \etal proposed a graphical model to find temporally consistent superpixels \cite{chang2013video}. Grundmann \etal proposed a hierarchical video segmentation technique, referred to as GBH,  based on appearance and regions graphs and also discussed its streaming and parallelizable version \cite{grundmann2010efficient}.  Xu \etal proposed a streaming video segmentation method based on GBH under a Markov assumption \cite{xu2012streaming}. Veksler \etal proposed a graph-cut based segmentation technique for image and video segmentaion \cite{veksler2010superpixels}. All of these methods assign labels to pixels and do not exploit dense trajectories and do not ensure a single label for a trajectory.

A somewhat related problem to ours is trajectory based motion and video segmentation. The goal of these methods is usually to find generic objects given a set of dense or sparse trajectories. Yan and Pollefeys segmented trajectories into articulated, rigid, non-rigid, degenerate and non-degenerate classes by finding a linear manifold embedding \cite{yan2006general}. Rao \etal. proposed a subspace clustering scheme to find out multiple moving objects in the video \cite{rao2008motion}. Ochs and Brox proposed a variational approach to obtain dense segmentation from sparse trajectories \cite{ochs2011object}.  Fragkiadaki \etal exploited the discontinuity in a trajectory embedding to segment out the objects \cite{fragkiadaki2012video}. Ochs \etal used long-trajectories and the affinities between them to segment different objects in the video \cite{ochs2014segmentation}. Keuper \etal. cast the motion segmentation as a minimum cost multicut problem \cite{keuper2015motion}. Wang \etal proposed a semi-supervised method to segment a foreground object from a video by clustering trajectories \cite{wang2017super}. They also coined the term super-trajectories for a trajectory cluster. In contrast to their work, we discuss the problem of trajectory clustering with the goal that a temporal slice of super-trajectory should be a super-pixel and the disconnectivity among the trajectories in the cluster should be minimized.

%=====================================================================================
\section{Trajectory Based Video Representation}
%\section{Method}
\subsection{Optical Flow to Trajectories}\label{sec:OF2Traj}

Optical flow provides a dense correspondence of pixels between two images. In order to exploit long-term temporal consistency, we convert the flow into long trajectories. Given forward and backward flow and edge images of $F$ frames, each of height $H$ and width $W$, the goal of this section is, to construct $T$ trajectories covering every pixel in the sequence. Each trajectory, $\mathbf{X}^i$, where $i \in \{1,\hdots,T\}$, is an $F\times1$ vector of tuples, each carrying 2D image coordinates and some of them may consist of missing values due to pixels' occlusion. $T~\gg~HW$, because of new scene entering into the field of view, and also for every occlusion, new pixels appear. We stack all the trajectories into a $F~\times~T$, sparse matrix, $\mathbf{X}$, which we estimate as the following.

The forward and the backward flows can be used to find current to previous and previous to current maps of pixel coordinates for the frame $f$, $\mathbf{X}_{f\_p}$ and $\mathbf{X}_{p\_f}$ respectively, where $p=f-1$. The composition of these maps gives the required dense trajectories. The key challenge in doing so is first finding out which pixels belong to a new scene so that the existing trajectories should be terminated and the new trajectories could be formed. Traditionally, the forward and backward optical flow consistency is used to find out new scene in the images\cite{sundaram2010dense} . This, however, results in breaking up good tracks and generating a large number of trajectories of short durations. We combine optical flow with color and edge boundary information and proposed a more robust criterion to find out new regions as follows.

We find $\mathcal{D}_{fO}$, $\mathcal{D}_{fC}$ and $\mathcal{D}_{fB}$ as $H\times W$ distance matrices for optical flow, color and edge boundary for the $f^\mathrm{th}$ frame, where  $\mathcal{D}_{fO}$ is the the Euclidean distance between $\mathbf{X}_{f\_p}$ and the inverse of $\mathbf{X}_{p\_f}$, $\mathcal{D}_{fC}$ is the Euclidean distance between 3-channel previous color image $\mathcal{C}_p$ and the warping of the $f^\mathrm{th}$ image to the previous image, $\mathcal{C}_{f\_p}$ using $\mathbf{X}_{f\_p}$ and $\mathcal{D}_{fB}$ is estimated from the edge boundary image of the previous frame $\mathcal{B}_p$ and the warped boundary image $\mathcal{B}_{f\_p}$ as follows,
\begin{equation}\label{Eq:Db}
\mathcal{D}_{fB} = \exp(\beta\max(\mathcal{B}_p, \mathcal{B}_{f\_p})),
\end{equation}
where both $\exp()$ and $\max()$ are element-wise functions and $\beta$ is a constant and was set equal to 4. We find the joint distance as follows,
\begin{equation}\label{Eq:Db}
\mathcal{D}_{f} = \left(\mathcal{D}_{fO} + \mathcal{D}_{fC}/\sigma \right)\circ \mathcal{D}_{fB},
\end{equation}
where $\sigma$ is a constant (we set this equal to 20) and $\circ$ denotes element-wise product. We define the optical flow at pixel $\mathbf{x}$ as inconsistent if $\mathcal{D}_{f}(\mathbf{x})\geq \gamma$, where $\gamma$ is a constant. Smaller $\gamma$ gives more trajectories of smaller duration and higher accuracy and vice-versa. We discuss the choice of $\gamma$ later in this section.

\begin{figure}
\begin{tabular}{cccc}
\includegraphics[width=\v]{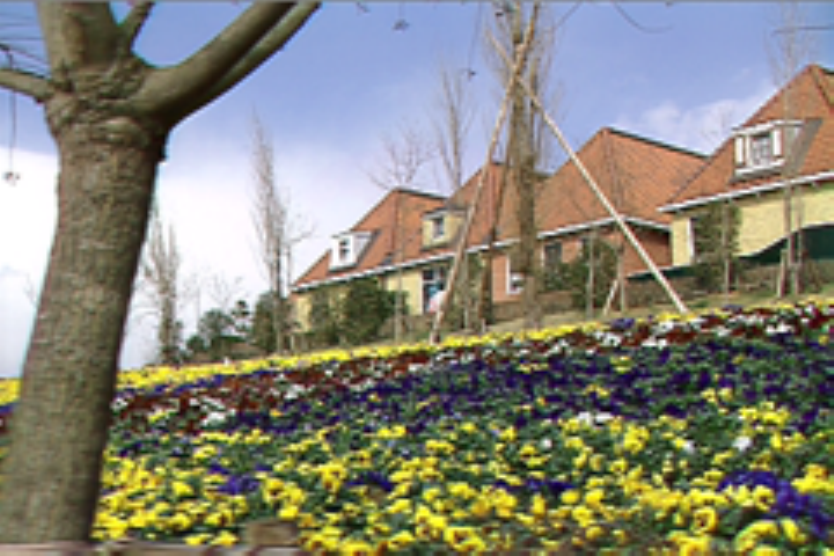}&
\includegraphics[width=\v]{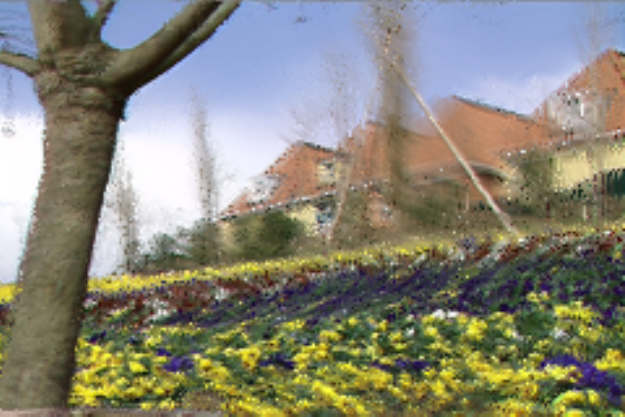}&
\includegraphics[width=\v]{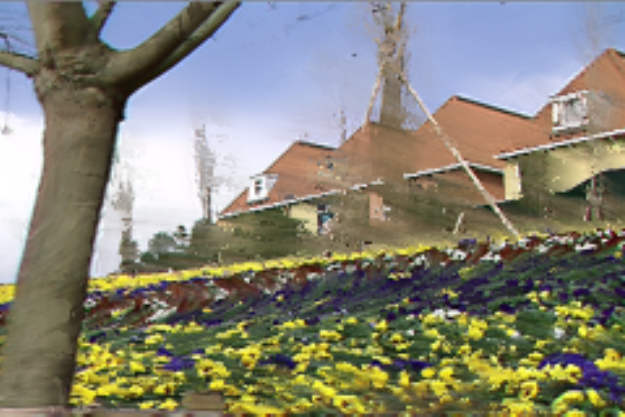}&
\includegraphics[width=\v]{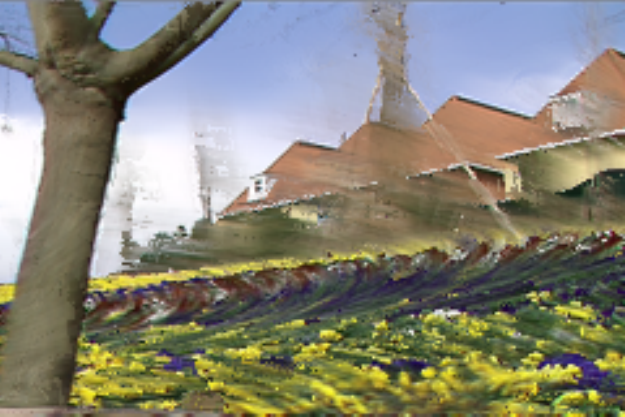}\\
\includegraphics[width=\v]{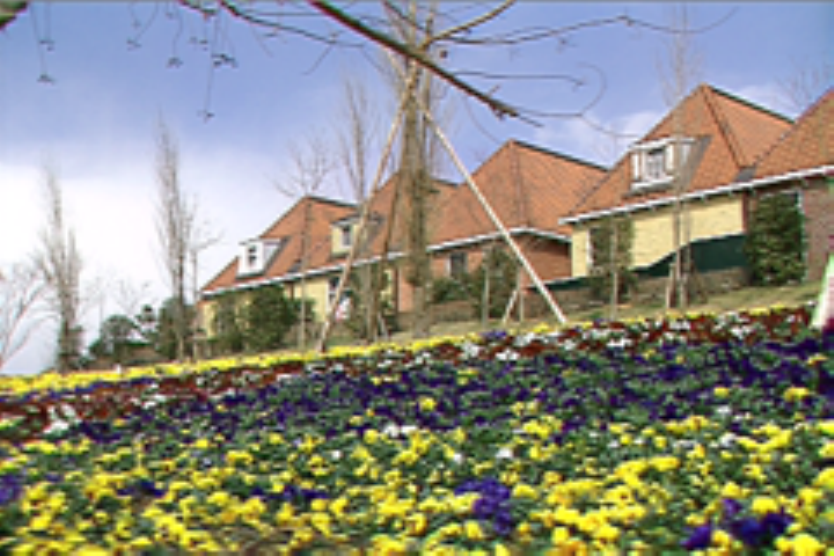}&
\includegraphics[width=\v]{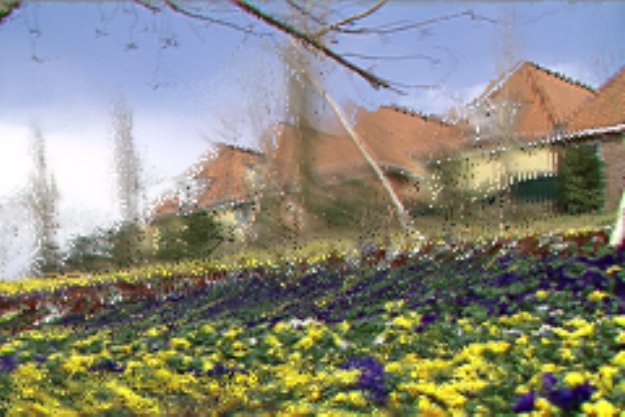}&
\includegraphics[width=\v]{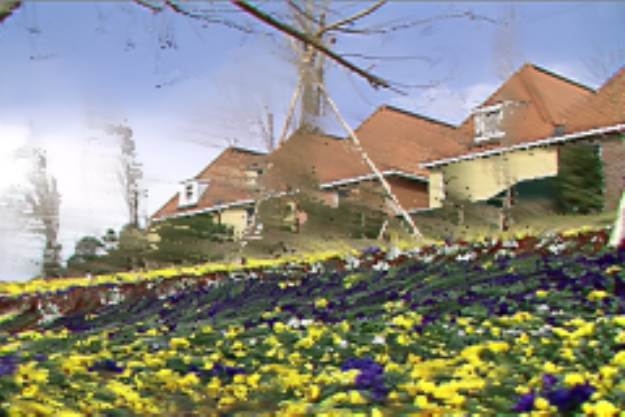}&
\includegraphics[width=\v]{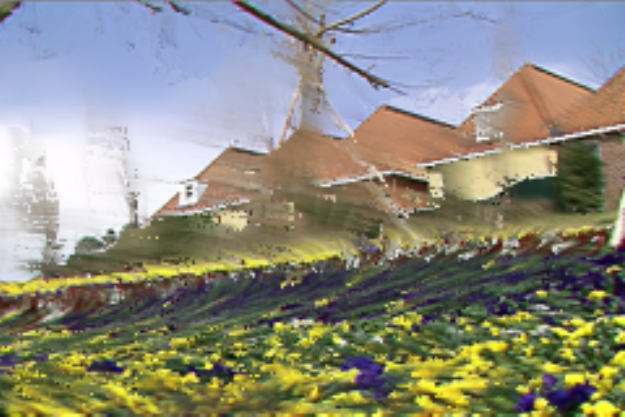}\\
\smaller{Original Images} & \smaller{OF-consistency} & \smaller{Proposed, $\gamma=0.9$} & \smaller{Proposed, $\gamma=1.5$}\
\end{tabular}
\caption{Qualitative comparison of trajectories estimation. We estimate trajectories using OF-consistency and the proposed method and then find the average color per trajectory and regenerate the two images with them and the corresponding trajectory coordinates. The proposed method with $\gamma=0.9$ gave roughly equal number of trajectories to OF-consistency, while significantly sharper images. With $\gamma=1.5$, it gives $40\%$ less trajectories while still maintaining the sharp edges in the images.}\label{fig:trajWarping}
\end{figure}

The area of inconsistent flow is considered as the occluded region and we set the corresponding flow as undefined. This helps us initialize new trajectories corresponding to the new regions. The goal of trajectory construction is to find, $\mathbf{X}_{f\_1}$, the mapping of every frame $f$ w.r.t the first frame. For the pixels not visible in the first frame, the mapping is defined w.r.t their first occurrence in a later frame. A row-wise stacking of $\mathbf{X}_{f\_1}$ for all $f$, gives the matrix $\mathbf{X}$

%The forward and the backward flows can be used to find current to previous and previous to current maps of pixel coordinates, $\mathbf{X}_{c\_p}$ and $\mathbf{X}_{p\_c}$ respectively, where $p=c-1$. We invert $\mathbf{X}_{p\_c}$ and match it with $\mathbf{X}_{c\_p}$  to find out the consistent flow [edit here!]. The area of inconsistent flow is considered as the occluded region and we set the corresponding flow as undefined. This helps us initialize new trajectories corresponding to the new regions. The goal of trajectory construction is to find, $\mathbf{X}_{c\_1}$, the mapping of every frame $c$ w.r.t the first frame. For the pixels not visible in the first frame, the mapping is defined w.r.t their first occurrence in a later frame. A row-wise stacking of  $\mathbf{X}_{c\_1}$ for all $c$, gives the matrix $\mathbf{X}$.

The first two rows in $\mathbf{X}$ would simply be $\mathbf{X}_{1\_1}$ and $\mathbf{X}_{2\_1}$, where $\mathbf{X}_{1\_1}$ represents an enumeration of the 2D coordinates of all the pixels in the first frame. The occluded region in the $2^\textrm{nd}$ frame would initialize new trajectories. The coordinates of the $j^\textrm{th}$ trajectory in the $3^\textrm{rd}$ frame, $\mathbf{x}_{j,3\_1}$ can be obtained from $\mathbf{x}_{j,2\_1}$ and $\mathbf{x}_{j,3\_2}$, as follows,

\begin{equation}\label{Eq:xj3}
\mathbf{x}_{j,3\_1} = \mathbf{X}_{3\_2}(\mathbf{x}_{j,2\_1}),
\end{equation}
where $\mathbf{X}_{f\_p}(\mathbf{x}_{j,p\_1})$ gives the mapped coordinates of the pixel location $\mathbf{x}_{j,p\_1}$ in $\mathbf{X}_{f\_p}$. In practice, since the flow would only give the mappings of discrete locations in the $2^\textrm{nd}$  frame, bilinear interpolation needs to be done to find the value $\mathbf{x}_{j,3\_2}$ at floating coordinates given by $\mathbf{x}_{j,2\_1}$. Please note that Equation \ref{Eq:xj3} is valid for both existing and new trajectories because we allow $\mathbf{x}_{j,2\_1}$ to also be a new point starting from the frame 2. Generalizing the above equation, $\mathbf{X}_{f\_1}$ can be obtained as the following recursive composition,
\begin{equation}
\mathbf{X}_{f\_1} = \mathbf{X}_{f\_p} \circ \mathbf{X}_{p\_1}.
\end{equation}
For every occluded region in the frame $f$, we initialize new trajectories and concatenate coordinates of the new region as additional columns in $\mathbf{X}_{f\_1}$. Hence all the pixels in the video are covered. To simplify the notation, we write $\mathbf{X}_{f\_1}$ as just $\mathbf{X}_f$ in the rest of the paper.

The goal of this paper is to cluster trajectories into $K$ classes. Let the $1\times T$ matrix $\mathbf{L}$ denotes the labels of the trajectories. By enforcing single output label for the entire duration of a trajectory, we not only reduce the number of unknowns but also explicitly enforce a long-term temporal consistency. Given the trajectory labels in $\mathbf{L}$, the pixel labels for the frame $f$,  $\mathcal{L}_f$ can be found as a $H\times W$ matrix using the corresponding trajectory coordinates in $\mathbf{X}_f$. We introduce a function $g()$ to denote this estimation of $\mathcal{L}_f$ as follows,
\begin{equation}\label{Eq:g}
\mathcal{L}_f = g(\mathbf{X}_{f},\mathbf{L}),
\end{equation}
where $g$ converts $\mathbf{X}_{f}$ into pixels with values taken from $\mathbf{L}$.

Analogous to the position matrix, $\mathbf{X}$, a $F\times T$ color matrix $\mathbf{C}$ can also be formed, where its $j^\textrm{th}$ column, $\mathbf{C}^j$ represents the 3D color values of the corresponding pixel locations in the trajectory, $\mathbf{X}^j$. This can be done using the following function
\begin{equation}\label{Eq:h}
\mathbf{C}_{f} = h(\mathbf{X}_{f}, \mathcal{C}_f),
\end{equation}
where $\mathcal{C}_f$ is the 3-channel color image and $\mathbf{C}_{f}$ represents the trajectory colors in the $f^\textit{th}$ frame and $h()$ plays an inverse role to $g()$. Similarly if edge boundaries for all the frames are given, then we can find an $F\times T$ matrix $\mathbf{B}$, consisting of the edge boundary values of the trajectories. Hence in the trajectory based video representation, certain features like color, edges, and positions can be described and compared at the trajectory level rather than the pixel level. 

In Fig \ref{fig:trajWarping} we give a qualitative comparison of the trajectories estimation using optical flow based forward-backward consistency check (OF-consistency) and the proposed method against $\gamma=0.9$ and $\gamma=1.5$. $\gamma=0.9$ gave the same number of trajectories as OF-consistency, whereas $\gamma=1.5$ gave roughly $40\%$ fewer trajectories. We find the average trajectory colors and then regenerate the frames using the formula, $\mathcal{C}_f = g(\mathbf{X}_{f},\mathbf{C})$. Better trajectories are expected to generate a sharper image back. The proposed method with $\gamma=0.9$ generates overall sharper images than OF-consistency with the same number of trajectories, while $\gamma=1.5$ generates fewer trajectories and still mostly sharp images. In our experiments we set $\gamma=1.5$.

The estimated trajectories are then used for clustering. Before describing the proposed method, we need to first define a few primitives of this representation.

%Analogous to the position matrix, $\mathbf{X}$, a $3F\times P$ color matrix $\mathbf{C}$ can be formed, where its $j^\textrm{th}$ column, $\mathbf{C}^j$ represents the color values of the corresponding pixel locations in the trajectory, $\mathbf{X}^j$. Thus the estimated trajectories provide an alternate video representation, where certain features like color or position can be described and compared at the trajectory level rather than pixel level. In addition, enforcing single output label for the whole trajectory not only reduces the number of unknowns but also explicitly enforces long-term temporal consistency. 
%
%Given the coordinates for the current frame, $\mathbf{X}_c$, the trajectory labels, $\mathbf{L}$ can be converted into a $H\times W$ matrix $\mathcal{L}_c$, representing the pixel labels of the frame, as the following
%\begin{equation}
%\mathcal{L}_c = g(\mathbf{X}_{c},L),
%\end{equation}
%where $g$ converts $\mathbf{X}_{c}$ into pixels with values taken from $\mathbf{L}$.
%The estimated trajectories are then used for clustering. Before describing the proposed method, we need to first define the trajectory primitives.

%=====================================================================================
\subsection{Trajectory Primitives}

Two trajectories are \textit{neighbours} if by rounding the coordinates, the corresponding pixels are  neighbours in at least one frame. We denote $N(\mathbf{X}^i)$ as the neighbours of the trajectory $\mathbf{X}^i$.

For a trajectory $\mathbf{X}^i$, we define its disconnectivity, as a $F\times1$ binary vector $\mathbf{d}$, w.r.t a set of trajectories, $\mathbf{S}_l = \{ \mathbf{X}^j \}$ with label $\mathbf{L}(\mathbf{X}^j )=l$, such that $\mathbf{d}$ is 1 at the frame $f$, if $\mathbf{X}^i$ is disconnected to $\mathbf{S}_l$ and 0 otherwise, i.e.
%Since two neighbouring trajectories may not be connected throughout the time, we define a measure of \textit{disconnectivity} between two trajectories. Consider a set of trajectories, $\mathbf{S}_l = \{ \mathbf{X}^j \}$ with label $\mathbf{L}(\mathbf{X}^j )=l$. We represent the disconnectivity of the trajectory $\mathbf{X}^i$ w.r.t  $\mathbf{S}_l$ as a binary vector, $\mathbf{d}\left(\mathbf{X}^i, \mathbf{S}_l\right)$ and define its value at the frame $f$ as
\begin{equation}
\mathbf{d}\left(\mathbf{X}^i, \mathbf{S}_l; f \right) = \begin{cases}
    1,& \text{if for all} ~\mathbf{X}^k \in N(\mathbf{X}^i), \mathbf{x}_{kf} \notin \{\mathbf{x}_{Sf} \}\\
    0,              & \text{otherwise}
\end{cases},
\end{equation}
where $\mathbf{x}_{kf}$ denotes the 2D coordinates at the frame $f$ of the trajectory $\mathbf{X}^k$ and $\{\mathbf{x}_{Sf} \}$, the set of 2D coordinates of $\mathbf{S}_l$ at the frame $f$. Similarly, we denote $\mathbf{v}_{j}$ as a binary visibility mask of $\mathbf{X}^j$ and define $\mathbf{v}_{S}$ as the visibility mask of $\mathbf{S}_l$ at the frame $f$ as following
\begin{equation}
\mathbf{v}_{S}(f) = \begin{cases}
    1,& \text{if }  \mathbf{v}_{j}(f)=1 ~\textrm{for~any} ~\mathbf{X}^j \in \mathbf{S}_l\\
    0,              & \text{otherwise}
\end{cases}. 
\end{equation}
Finally we define the cost of connectivity of $\mathbf{X}^i$ and $\mathbf{S}^l$ if $\mathbf{L}(\mathbf{X}^i)=l$, by simply counting the frames in which $\mathbf{S}_l$ was visible but $\mathbf{X}^i$ disconnected to $\mathbf{S}_l$, i.e.
\begin{equation}\label{Eq:cost}
\mathrm{cost}(\mathbf{X}^i, \mathbf{S}_l) = \sum_{f=1}^F \mathbf{d}\left(\mathbf{X}^i, \mathbf{S}_l; f \right) \& \mathbf{v}_{S}(f),
\end{equation}
where $\&$ is the boolean AND operator. Finally we say that the group  $\mathbf{S}_l\cup \mathbf{X}^i$ is fully connected if $\mathbf{S}_l$ consists of fully connected trajectories, i.e. $\mathrm{cost}(\mathbf{X}^j, {\mathbf{S}_l\setminus\mathbf{X}^j})=0$, for all $\mathbf{X}^j \in \mathbf{S}_l$,  and $\mathrm{cost}(\mathbf{X}^i, \mathbf{S}^l) = 0$.

We also need to define the \textit{energy} function we want to minimize for super-trajectory clustering. Traditionally, this energy consists of color and position based terms and can be defined for a trajectory $\mathbf{X}^i$, as follows
\begin{equation}\label{Eq:tdis1}
E_{ik}= E_s(i,k) + E_{c}(i,k),
\end{equation}
where $E_{s}(i,k)$ is proportional to the squared Euclidean distance between the frame-level 2D coordinates of some center trajectory $\mathbf{X}^k$ and $\mathbf{X}^i$, averaged over the visible frames and $E_{c}(i,k)$ is an analogous energy term for the color. We modify the energy by also considering the already labeled neighbouring trajectories, $\mathbf{X}^j$ of $\mathbf{X}^i$ as follows
\begin{equation}\label{Eq:tdis}
E(i)= E_{ik} + \min_{j}E_b(i,j),
%E(i)= E_{ik} + \min_{j}\left(\left(E_c(i,j) + E_s(i,j)\right)E_b(i,j)\right),
\end{equation}
where  $E_b = \frac{1}{a}\sum_{f }\exp(\beta \max(b_{if}, b_{jf})),$
and $b_{if}$ denotes the edge boundary of the $i^\textrm{th}$ trajectory in the frame $f$ and $\beta$ is a constant, the summation is over the frames where both the trajectories were visible and $a$ is the number of these frames. Since minimzing the sum of the energies for all trajectories is a difficult optimization, we instead approximate it with a greedy optimization by selecting and labeling the trajectory with the least energy as discussed in the next section.

%Please note that for a single  $\mathbf{X}^i$ there should be multiple $E(i,j)$ because of multiple neighbours. Since there might be multiple $\mathbf{X}^j$s for a single $\mathbf{X}^i$, 

%The average superpixel diameter, $s$ and the spatial regularization, $m$ are the same parameters used in SLIC \cite{achanta2012slic} and are given by the user. 

\section{Super-Trajectory Representation}
We propose an iterative procedure for super-trajectory clustering, where in each iteration, we adapt SNIC \cite{achanta2017snic} algorithm for trajectories and improve the connectivity until convergence. Once the trajectory labels are found, the pixel labels of the frame $f$, $\mathcal{L}_f$ are found, using Equation \ref{Eq:g}. Here we describe different parts of the algorithm.
%=====================================================================================
%\subsection{Trajectory SNIC}
\subsection{The Core Algorithm}

\begin{algorithm}[t]
\SetKwInOut{Input}{input}
\SetKwInOut{Output}{output}
\Input{$\mathbf{X}, \mathbf{C}, s, m$}
\Output{$\mathbf{L}$}
Initialize cluster centers \;
$cnt \gets K; \qquad $  \Comment{\# of Clusters}\
$pcnt \gets \infty$\;
$iter \gets 0$\;
Set $\mathbf{L}^*$ for the cluster centers and the $\textrm{rest} =0$\;
\While{$cnt<pcnt~\&~cnt>0~\&~iter<maxIter$}{
	$\mathbf{L} \gets$ TNIC($\mathbf{X}, \mathbf{C}, \mathbf{L}^*, s, m$)\;
	\uIf{$iter$ is $0$}{
	Find the largest subset of fully connected trajectories and set $\mathbf{L}^*$\;
	}
	\Else{
	\For{$k\gets1$ \KwTo $K$}{
		Find the set, $\mathbf{S}_k = \{ \mathbf{X}^i\}$ with $\mathbf{L}^*[\mathbf{X}^i]=k$\;
		\For{$\mathbf{X}^j \in N(\mathbf{S}_k$) with $\mathbf{L}^*[\mathbf{X}^j]=0$ \& $\mathbf{L}[\mathbf{X}^j]=k$}{
			\If{$\mathrm{cost}(\mathbf{X}^j, \mathbf{S}_k)$ is $0$}{
				$\mathbf{L}^*[\mathbf{X}^j]=k$\;
				$\mathbf{S}_k \gets \mathbf{S}_k \cup \mathbf{X}^j $\;
			}
		}
	}
	}
	$cnt \gets $ number of nonzeros in $\mathbf{L}^*$\;
	$pcnt \gets cnt$\;
	$iter = iter + 1$\;
}
$\mathbf{L} \gets$ PostProcess($\mathbf{X}, \mathbf{L}$)\;
\caption{Super-Trajectory Clustering}\label{alg:STrj}
\end{algorithm}
%Then we find the cluster labels by assigning a trajectory to the nearest center in terms of the distance metric, Equation \ref{eq:tdis}. The estimated lables update the cluster centers and both the steps are repeated for a small number of iterations. After this, we run connected component labeling and grow the set of fully connected trajectories, i.e. ones that give zero connectivity cost, Equation \ref{Eq:cost}, for each cluster label. The largest subset is then used to initialize Algorithm \ref{alg:STrj}.

The proposed iterative method is given in Algorithm \ref{alg:STrj}, where each iteration reduces the number of disconnected trajectories and calls TNIC, the adaptation of SNIC for trajectories, given in the Algorithm \ref{alg:TNIC}, as a sub-routine. TNIC starts with a labeling of subset of fully connected trajectories and uses them to find cluster centers (line 2). It finds the distances of neighbours of the unlabeled trajectories and pushes them in a queue (lines 3-7). Then in the while loop, the trajectory corresponding to the smallest distance popes out, gets its label and updates the cluster centers (lines 12-15). After this the unlabeled neighbours of the trajectory are added to the queue (lines 16-19). The loop terminates when the queue is empty.

Algorithm \ref{alg:STrj} starts by finding a few trajectories as the seed cluster centers. $\mathbf{L}^*$ stores the labels of fully connected clusters and grows in each iteration. Initially only the seed cluster centers are labeled and the rest are labelled 0 (line 5). Each iteration of the while loop finds intermediate labeling $\mathbf{L}$ using TNIC (line 7). For the first iteration, we find the largest subset of fully connected trajectories using connected component labeling and set $\mathbf{L}^*$ accordingly (line 9). For the remaining iterations, we simply grow $\mathbf{L}^*$ if the corresponding label in $\mathbf{L}$ makes them fully connected with the existing fully connected trajectories (line 11-19). The loop terminates when the disconnected trajectories cannot be reduced anymore. A post-processing is done to get the final labeling $\mathbf{L}$ (line 25).
 
The initialization and the post-processing are discussed in the following sections.
\begin{algorithm}[t]
\SetKwInOut{Input}{input}
\SetKwInOut{Output}{output}
\Input{$\mathbf{X}, \mathbf{C}, \mathbf{L}_c, s, m$}
\Output{$\mathbf{L}$}
%\SetAlgoLined
%\KwResult{Write here the result }
 %initialization\;
 
 \For{labels: $k\gets1$ \KwTo $K$}{	
   Find $\mathbf{X}^k$ and $\mathbf{C}^k$, the centroid of the sets, $\{ \mathbf{X}^i\}$ and $\{ \mathbf{C}^i\}$ with $\mathbf{L}_c[\mathbf{X}^i]=k$ respectively\;
    %Find $C[k] = ( \mathbf{X}^k, \mathbf{C}^k)$, where $\mathbf{X}^k$ and $\mathbf{C}^k$ are the centroid of the sets, $\{ \mathbf{X}^i\}$ and $\{ \mathbf{C}^i\}$, with $\mathbf{L}_c[\mathbf{X}^i]=k$\;
    Find the unlabeled neighbours, $N(\{ \mathbf{X}^i\})$\;
    \For{$\mathbf{X}^j \in N(\{ \mathbf{X}^i\})$}{
    Find $E_{i}$ from Equation \ref{Eq:tdis}\;
    $e_j \gets (\mathbf{X}^j, \mathbf{C}^j, k, E_{i})$\;
    Push $e_j$ on priority queue $Q$\;
    }
    }
    $\mathbf{L} \gets \mathbf{L}^*$\;
 \While{$Q$ is not empty}{
  Pop $Q$ to get $e_j$\;
  \If{$\mathbf{L}[\mathbf{X}^j]$ is 0}{
  	$\mathbf{L}[\mathbf{X}^j] = k_j$\;
	Update centroid $C[k_j]$ online with $\mathbf{X}^j$ and $\mathbf{C}^j$\;
	\For{$\mathbf{X}^i \in N(\mathbf{X}^j)$ with $\mathbf{L}[\mathbf{X}^i]=0$}{
		$e_i \gets (\mathbf{X}^i, \mathbf{C}^i, k_j, E_{j})$\;
    		Push $e_i$ on priority queue $Q$\;
	}
  }
 }
 \caption{TNIC}\label{alg:TNIC}
\end{algorithm}
%=====================================================================================
\subsection{Initialization}
Ideally the seed cluster centers should be evenly separated, ensuring that all the trajectories lie within a circle of diameter roughly equal to the required average diameter of the superpixels $s$. For pixels as input, this can be trivially done by initializing seeds on a grid. For trajectories as input, each covering multiple pixels, this, however, is nontrivial and is discussed as the following.

%Ideally the seed cluster centers should be evenly separated. For pixels as input, this can be trivially done by initializing seeds on a grid. For trajectories as input, each covering multiple pixels, this, however, is nontrivial and is discussed as the following.

The seed cluster centers for the trajectories corresponding to the pixels in the $1^\textit{st}$ frame are simply initialize along a grid, with the spacing equal to $s$. This leaves the trajectories starting from the $2^\textit{nd}$ frame or later uncovered (i.e., not dealt with). Then we find the seeds for the uncovered trajectories ending at the last frame. After this we recursively find the middle frame among the previously considered frames, so that all the evenly spaced frames in time are considered or there are not enough uncovered trajectories left in a spatial window of size $s\times s$.

Let the $1\times T$ binary vector, $\mathbf{u}$ represent the uncovered trajectories after the selection of seeds from a frame. After the $1^\textit{st}$ frame, the uncovered trajectories would be $\mathbf{u} = h(\mathbf{X}_1, \mathbf{1})$\footnote{function $h()$, previously defined for color vectors, is now being used for binary vectors.}, where $\mathbf{1}$ is a $H\times W$ matrix of 1s. To find the new seeds in a frame $f$, we first find the uncovered pixels in the corresponding frame as
\begin{equation}\label{eq:uf}
\mathcal{U}_f = g(\mathbf{X}_{f}, \mathbf{u}).
\end{equation}
We convolve $H\times W$ matrix, $\mathcal{U}_f$ with an $s\times s$ template of ones. The pixel location at the maximum in the convolution, if greater than a threshold $th$, gives the corresponding trajectory as the next seed. If the maximum is less than $th$ then we do not look for any more seeds in the frame. Otherwise, we set the $s\times s$ window centered at the selected pixel equal to 0 in $\mathcal{U}_F$ and again find the location corresponding to the maximum sum in a $s\times s$ window and repeat. Once done with a frame $f$, the remaining uncovered trajectories are found as
\begin{equation}\label{eq:u}
\mathbf{u} = h(\mathbf{X}_{f}, \mathcal{U}_f),
\end{equation}

Repeated applications of equations \ref{eq:uf} and \ref{eq:u} for the selected frames ensures coverage of all the frames and the selected seeds are used to initialize the cluster centers. 

%=====================================================================================

\begin{algorithm}
\SetKwInOut{Input}{input}
\SetKwInOut{Output}{output}
\Input{$\mathbf{X}, \hat{\mathbf{L}}$}
\Output{$\mathbf{L}^*$\qquad\Comment{dimension: $1\times T$}}
\For{$f\gets1$ \KwTo $F$}{
	$\mathcal{\hat{L}}_f = g(\mathbf{X}_f, \hat{\mathbf{L}})$;	\qquad ~ \Comment{$\mathcal{\hat{L}}_f$ is $H\times W$}
	Get $\bar{\mathcal{L}}_f$ from $\mathcal{\hat{L}}_f$ by filtering small label regions\;
	$\bar{\mathbf{L}}_f = h(\mathbf{X}_f, \bar{\mathcal{L}}_f)$;  \qquad \Comment{$f^\textit{th}$ row of $\bar{\mathbf{L}}_{F\times T}$}
}
Find the trajectories that do not get multiple labels in $\bar{\mathbf{L}}$ to initialize $\mathbf{L}^*$ and set the rest = 0\;
\For{$k\gets1$ \KwTo $K$}{
		Find the set, $\mathbf{S}_k = \{ \mathbf{X}^i\}$ with $\mathbf{L}^*[\mathbf{X}^i]=k$\;
		\For{$\mathbf{X}^j \in N(\mathbf{S}_k$) with $\mathbf{L}^*[\mathbf{X}^j] =0$}{
			$\forall k_n \in \mathbf{L}^*[N(\mathbf{X}^j)]$\\
			$\mathbf{L}^*[\mathbf{X}^j ] = \argmin_{k_n}{\mathrm{cost}(\mathbf{X}^j, \mathbf{S}_{kn})}$
		}
	}
\caption{PostProcess}\label{alg:postprocess}
\end{algorithm}

\subsection{Post-Processing}
The post-processing (given in Algorithm \ref{alg:postprocess}) starts with a loop for all the frames and converts the trajectory labels into a $H\times W$ matrix, consisting of pixel labels for the $f^\textit{th}$ frame (line 1-2). Then it filters outs the small isolated label regions to get $\bar{L}_f$, using the standard connected component labeling based procedure (line 3). $\bar{L}_f$ is converted back into trajectory labeling  to get the $f^\textit{th}$ row, $\mathbf{L}_f$ in the matrix $\bar{\mathbf{L}}$ (line 4).  The trajectories that do not get multiple labels in $\bar{\mathbf{L}}$ are taken as clean and the corresponding labels are assigned to $\mathbf{L}^*$ (line 6). Each remaining trajectory gets the label of its neighbour that gives the least connectivity cost (lines 7-12). 

\begin{figure}
\begin{tabular}{cccc}
\includegraphics[width=\v]{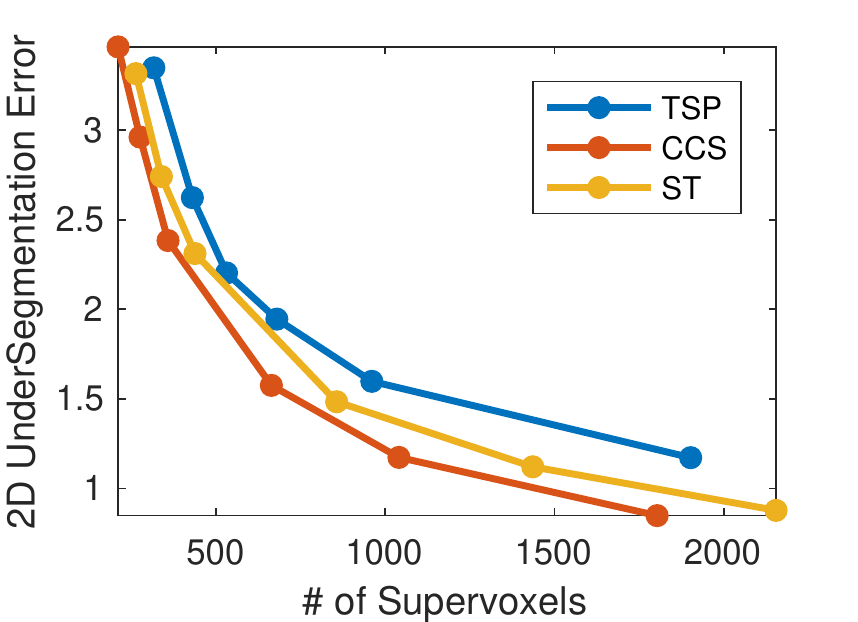}\label{fig:UE2D}&
\includegraphics[width=\v]{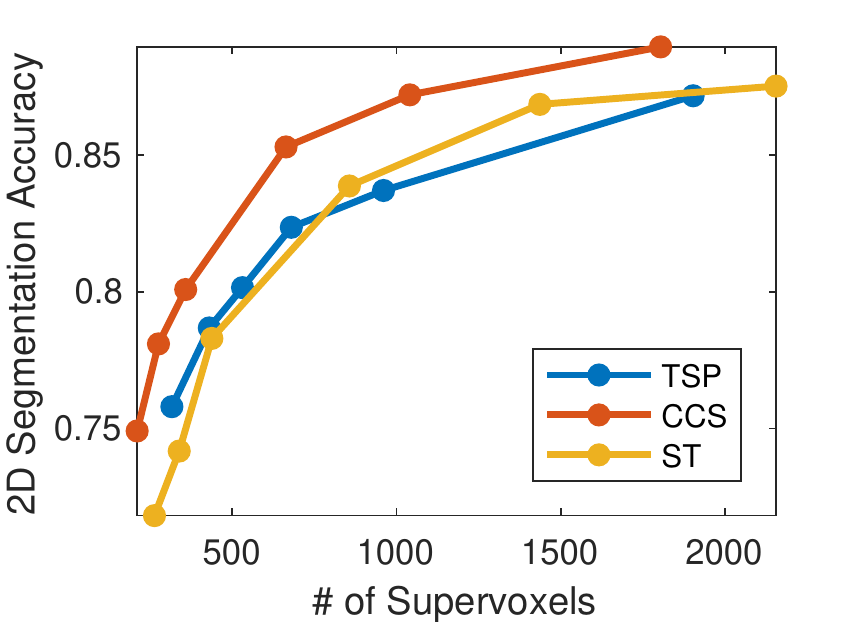}\label{fig:SA2D}&
\includegraphics[width=\v]{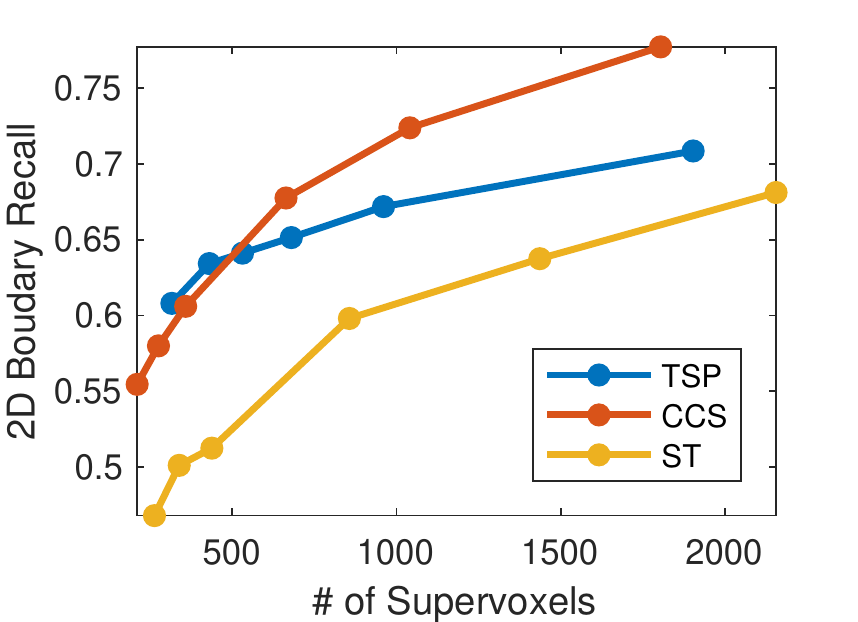}\label{fig:BR2D}&
\includegraphics[width=\v]{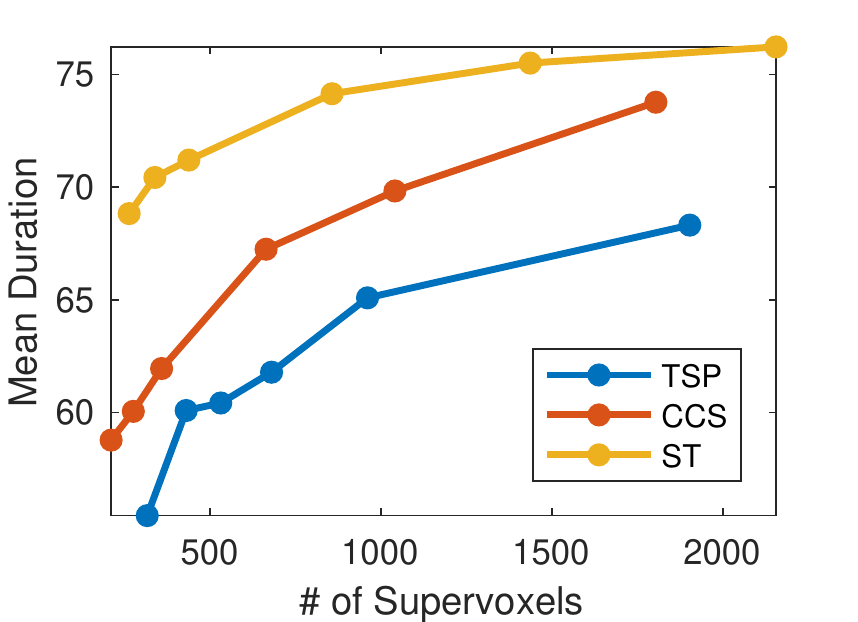}\label{fig:duration}\\
\smaller{UE2D $\downarrow$} & \smaller{SA2D $\uparrow$} & \smaller{BR2D $\uparrow$} & \smaller{Mean Duration $\uparrow$}\\
\includegraphics[width=\v]{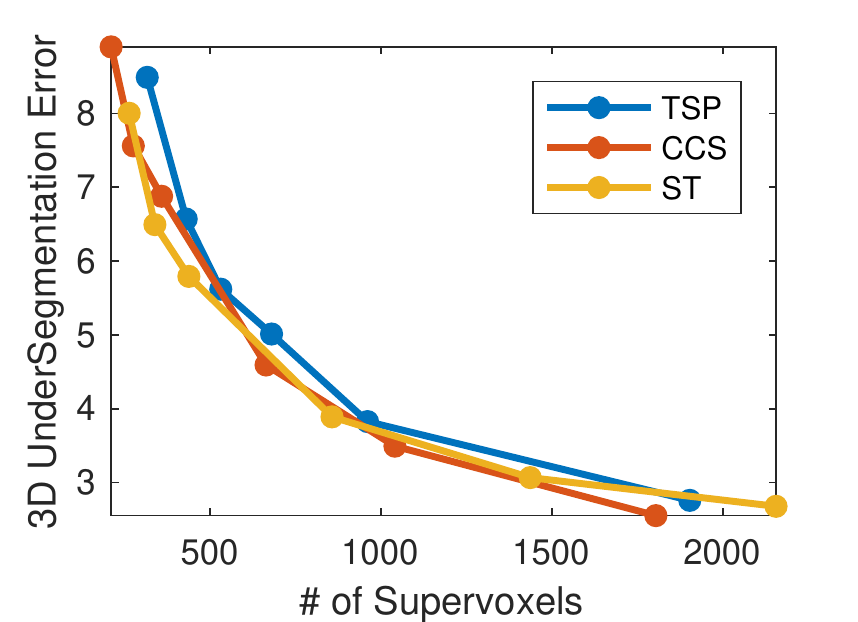}\label{fig:UE3D}&
\includegraphics[width=\v]{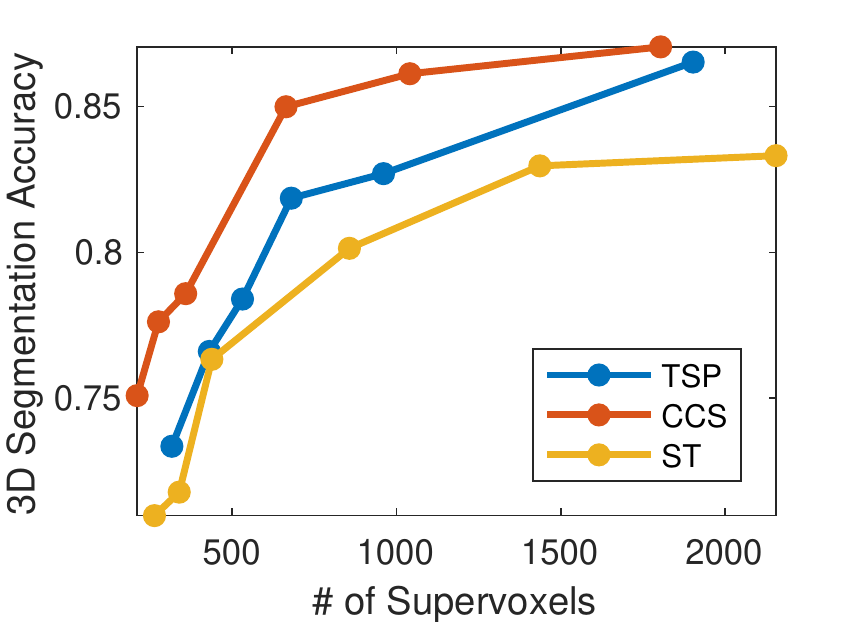}\label{fig:SA3D}&
\includegraphics[width=\v]{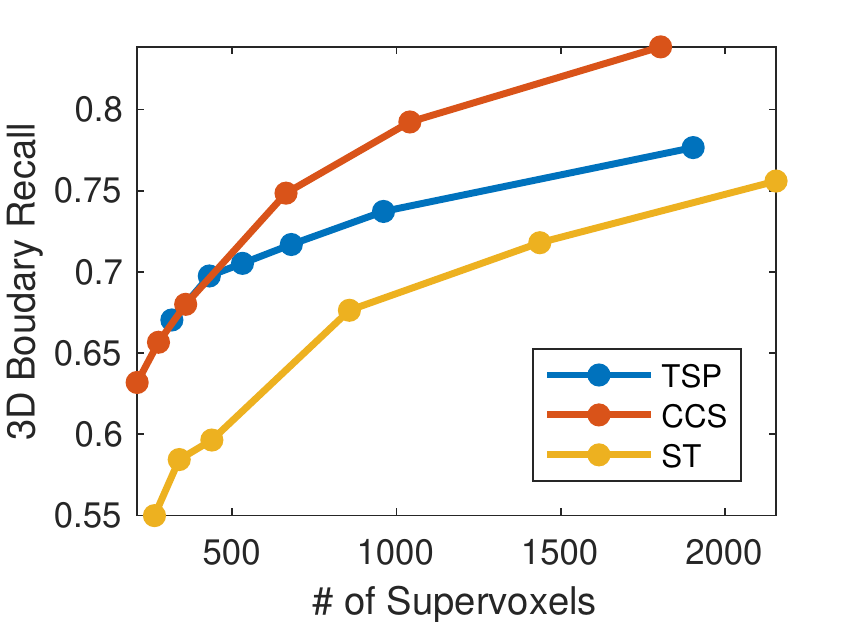}\label{fig:BR3D}&
\includegraphics[width=\v]{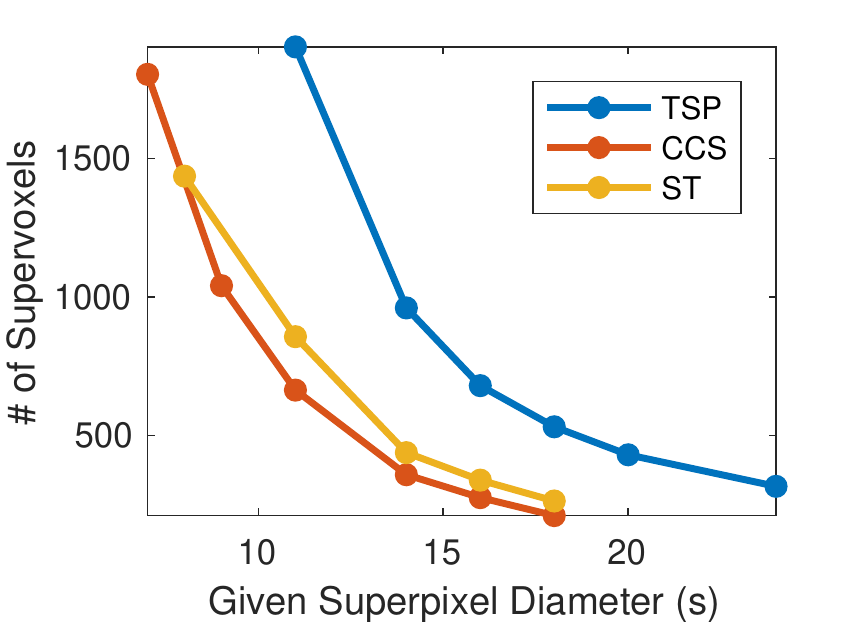}\label{fig:nSpxl}\\
\smaller{UE3D $\downarrow$} & \smaller{SA3D $\uparrow$} & \smaller{BR3D $\uparrow$} & \smaller{$\#$ of Supervoxels $\downarrow$}\
\end{tabular}
\caption{Quantitative evaluation of temporal superpixel algorithms versus super-trajectories (ST) on Chen dataset \cite{chen2010propagating} ($\downarrow$ means lower the better, whereas $\uparrow$ means the opposite). ST gives fewer superpixels of longer duration than CCS \cite{lee2017contour} and TSP \cite{chang2013video} against the number of supervoxels. Our under-segmentaion error is slightly better whereas segmentation accuracy and boundary recall are a little worse. Slight degradation in these metrics is the cost we need to pay to retain the full trajectories in the same cluster.}\label{fig:quantitative}
\end{figure}

\section{Experiments}

\begin{figure}[t]
\centering
\includegraphics[width=\w]{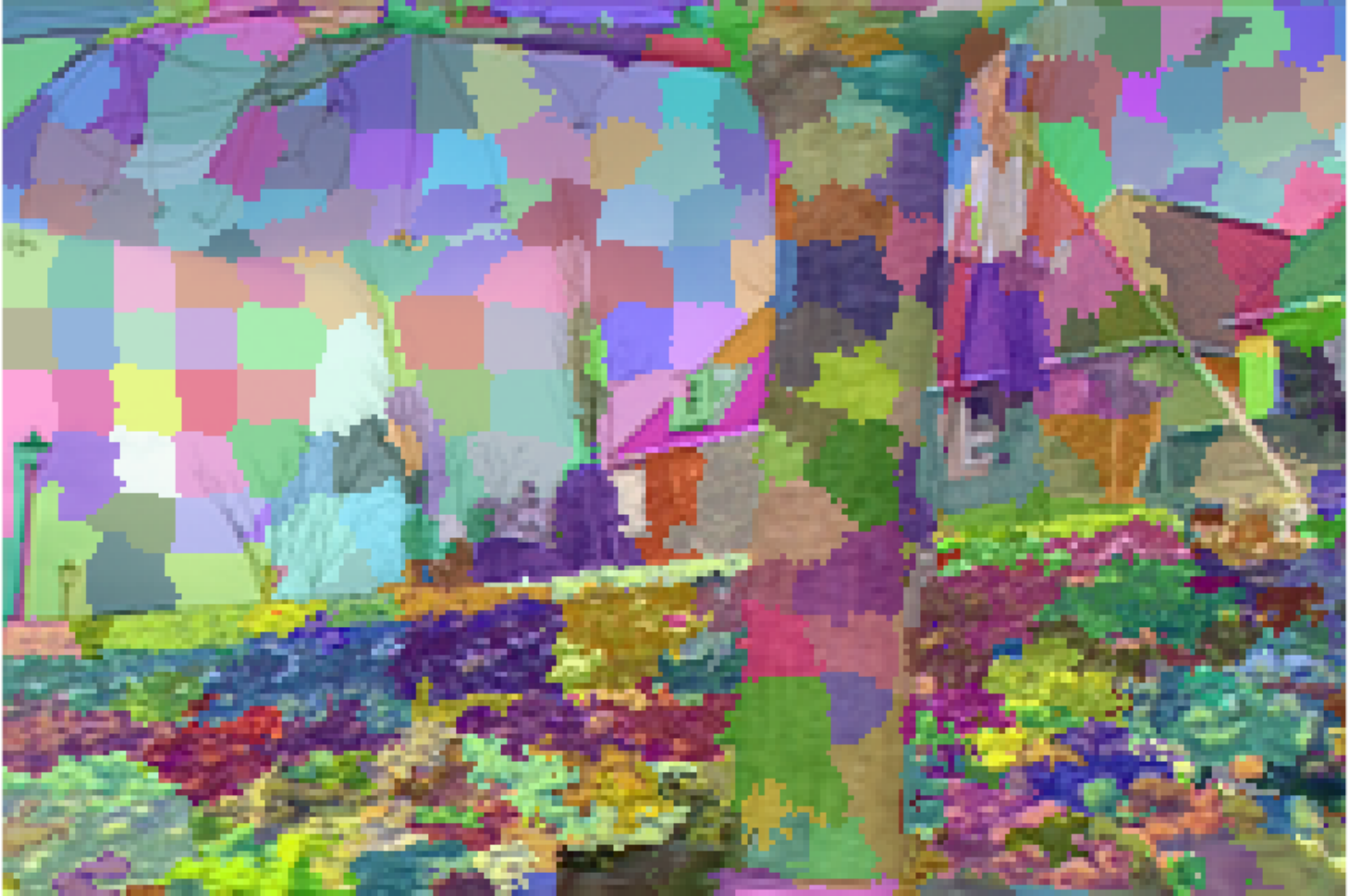}
\includegraphics[width=\w]{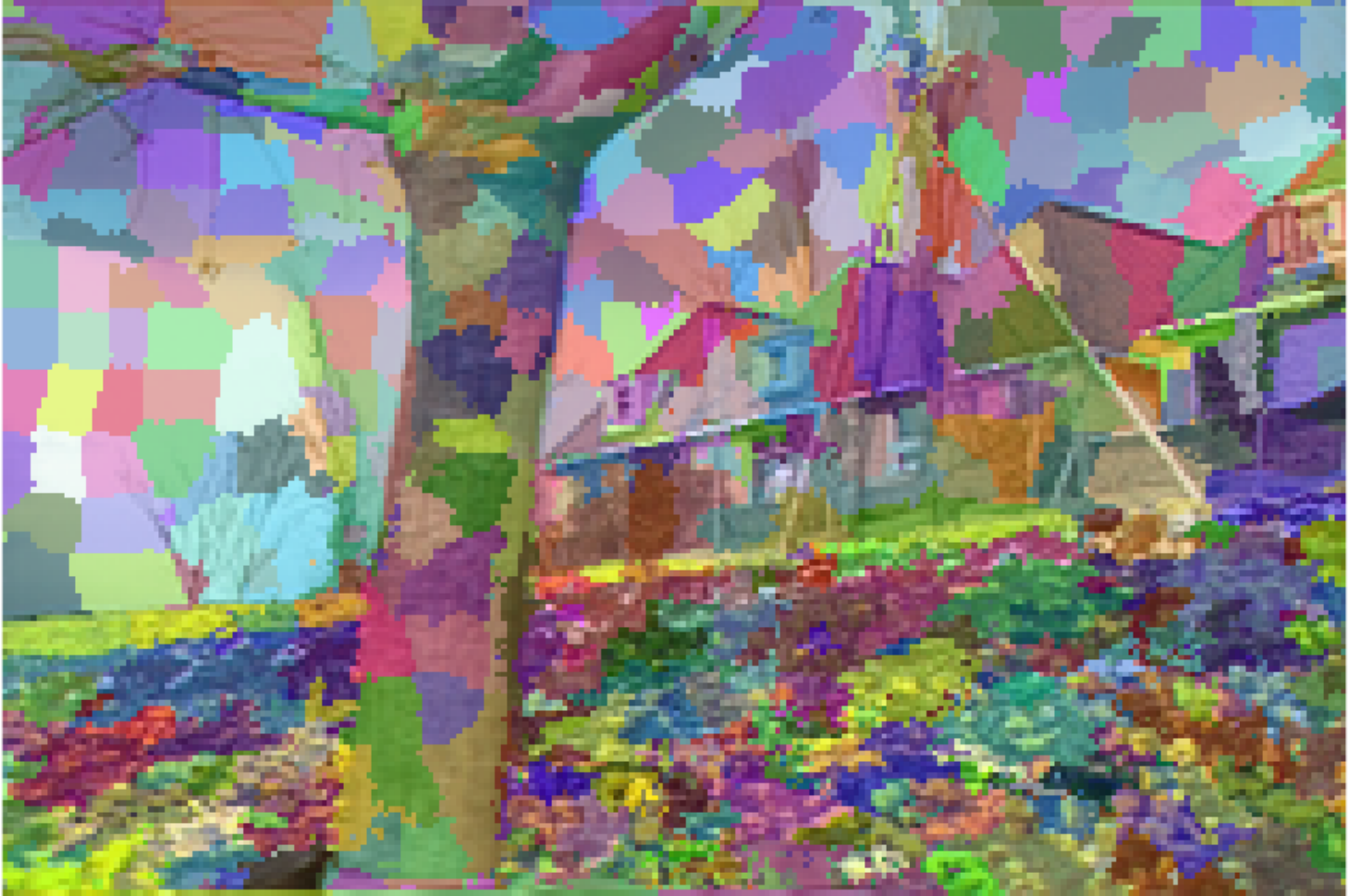}
\includegraphics[width=\w]{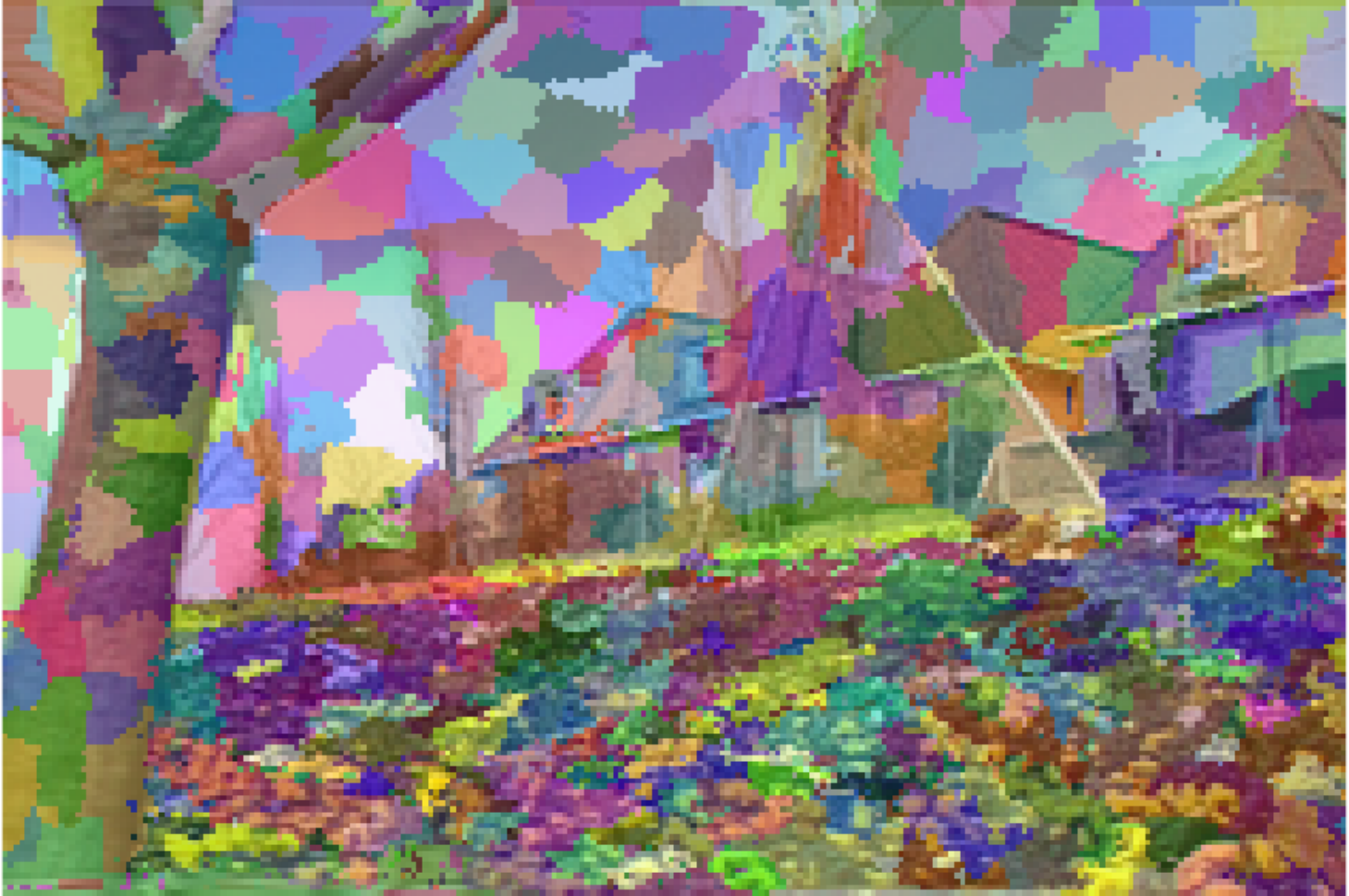}
\includegraphics[width=\w]{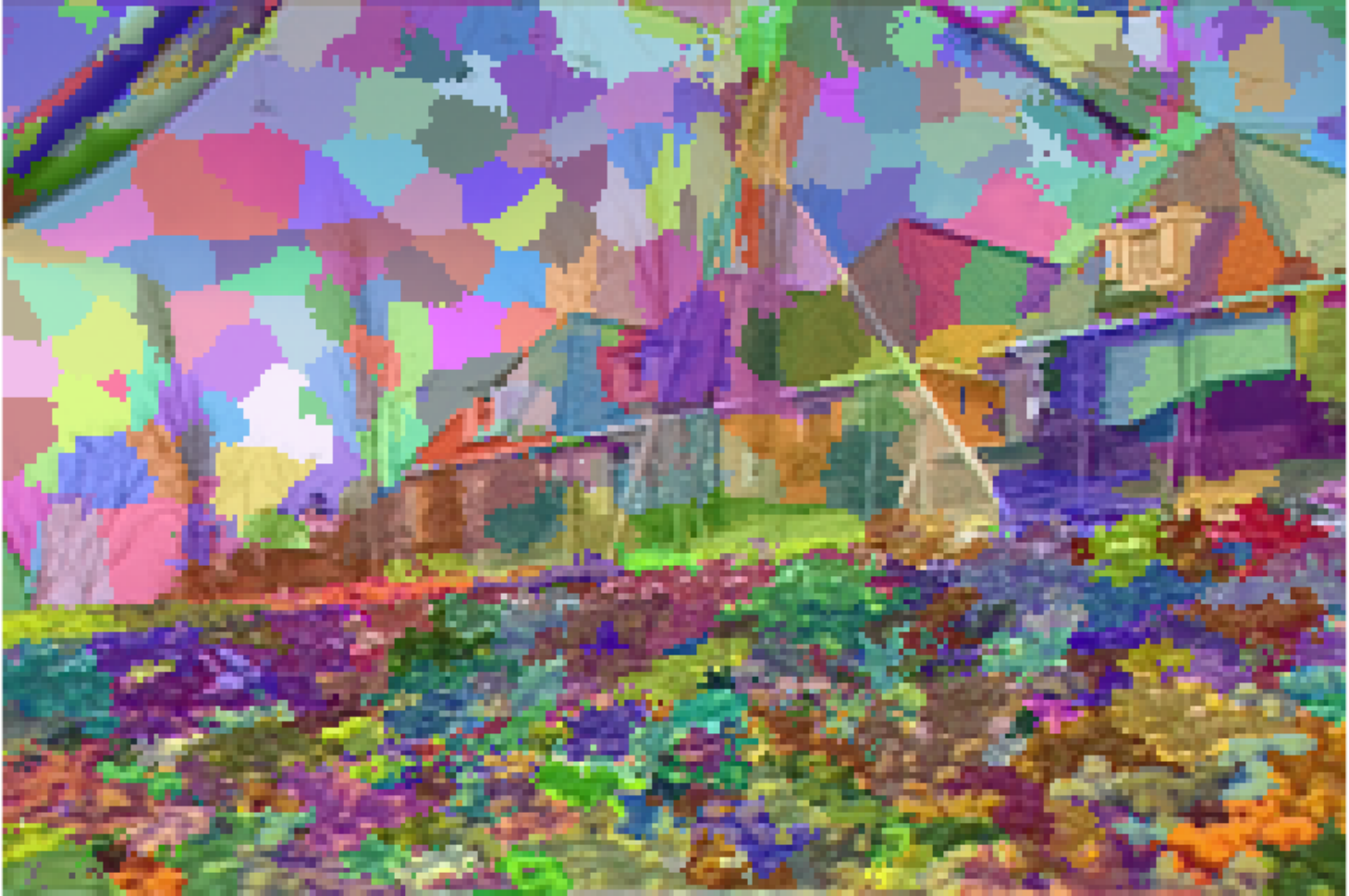}
\includegraphics[width=\w]{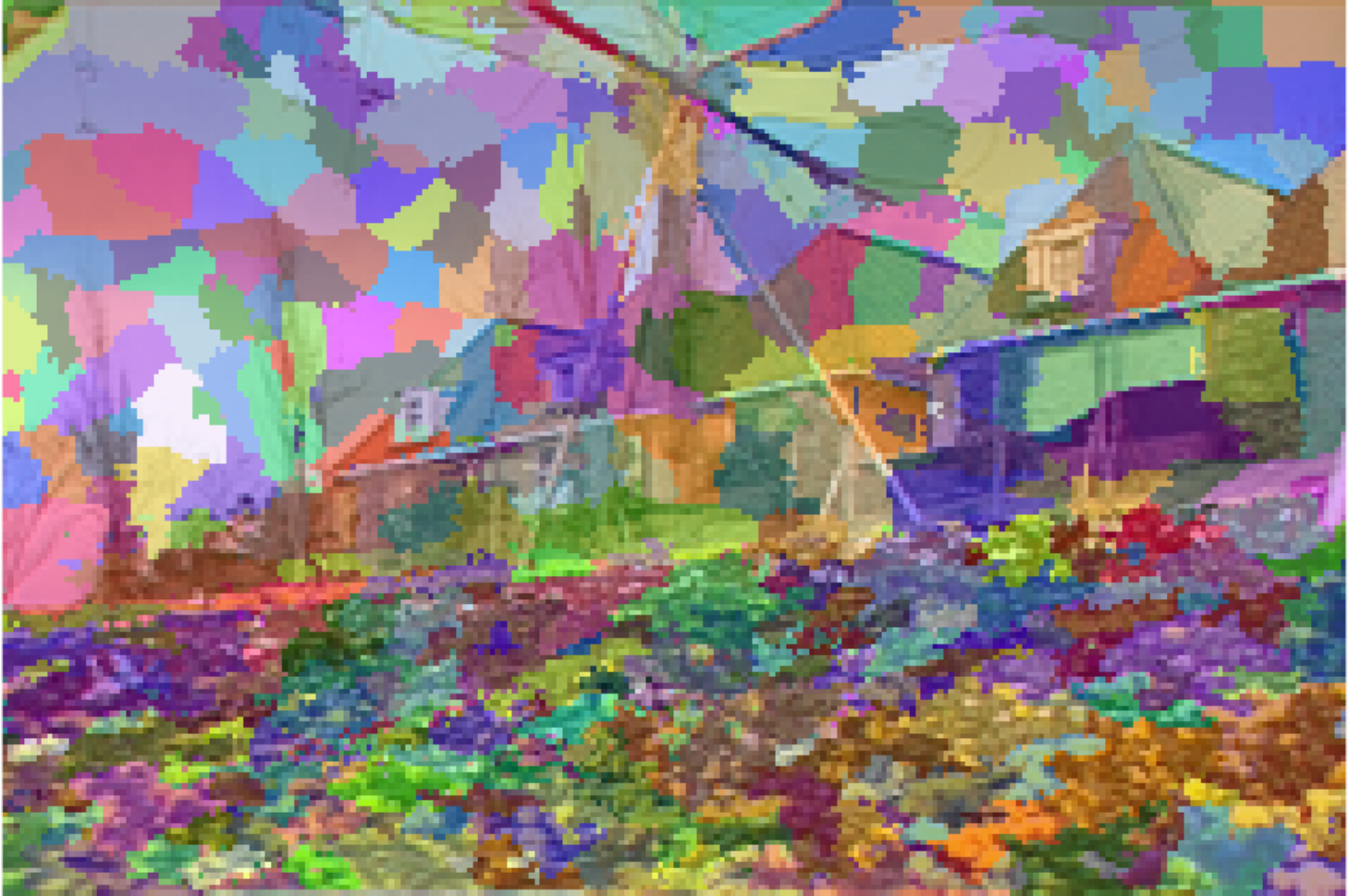}

\includegraphics[width=\w]{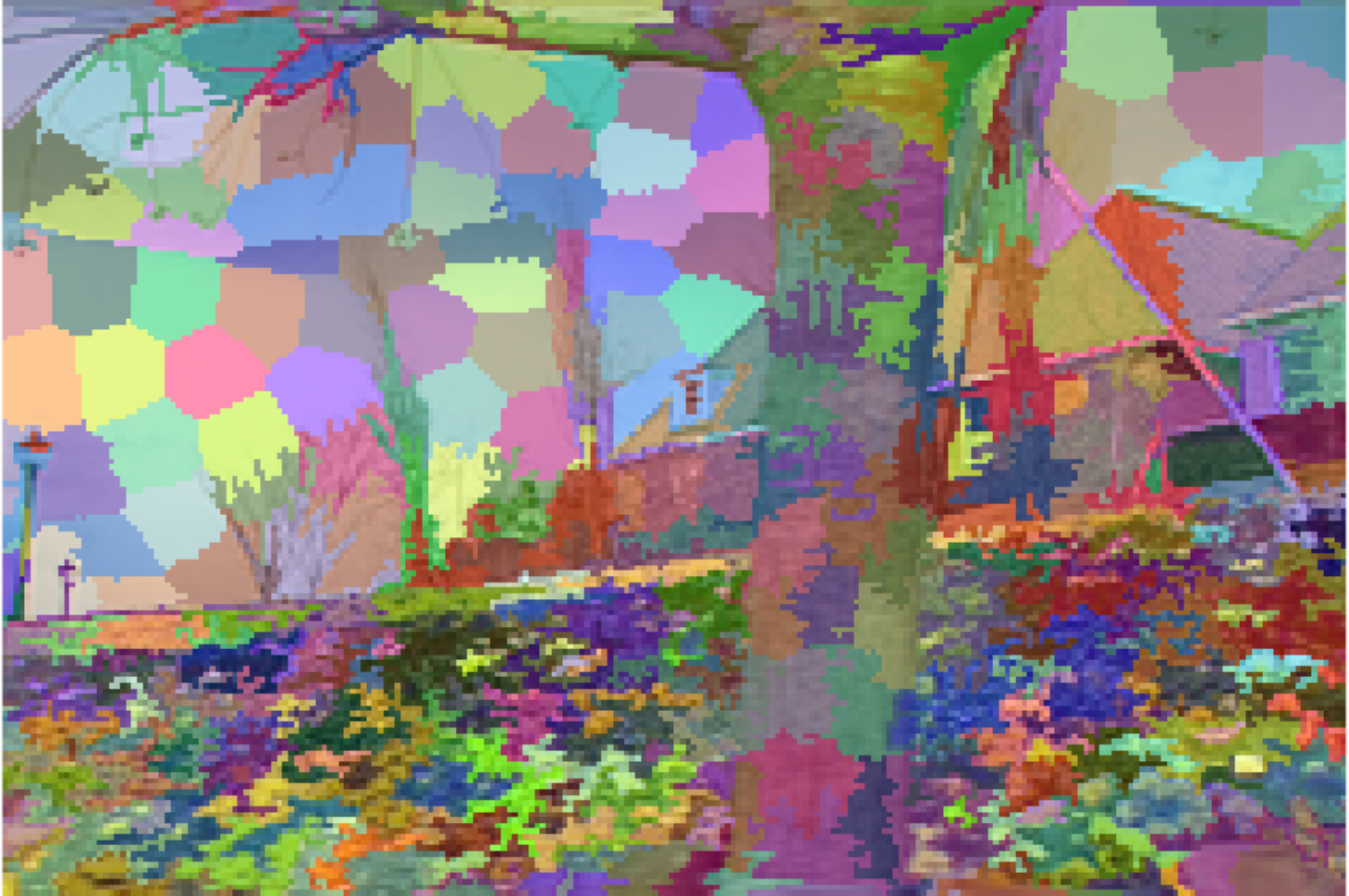}
\includegraphics[width=\w]{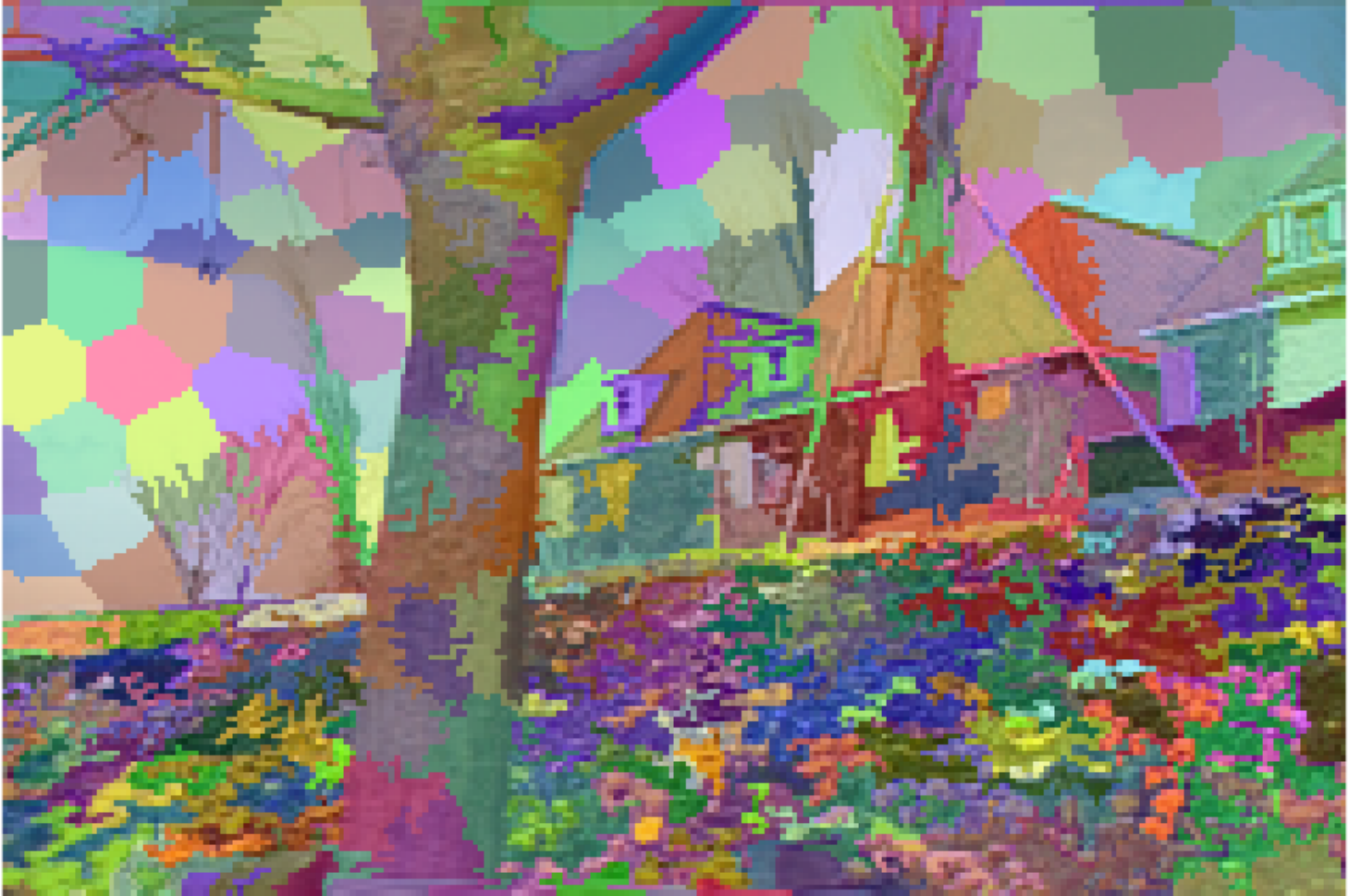}
\includegraphics[width=\w]{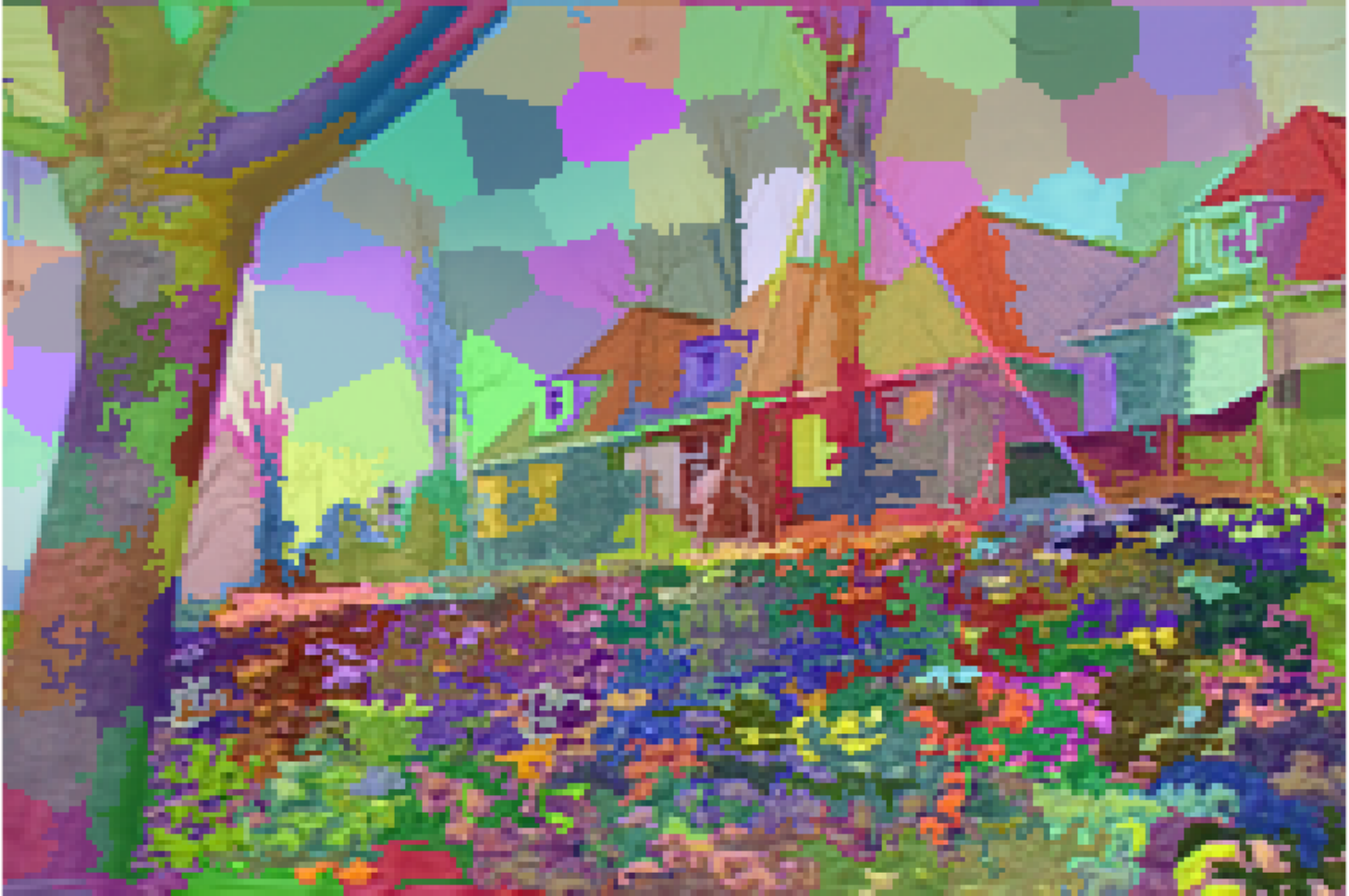}
\includegraphics[width=\w]{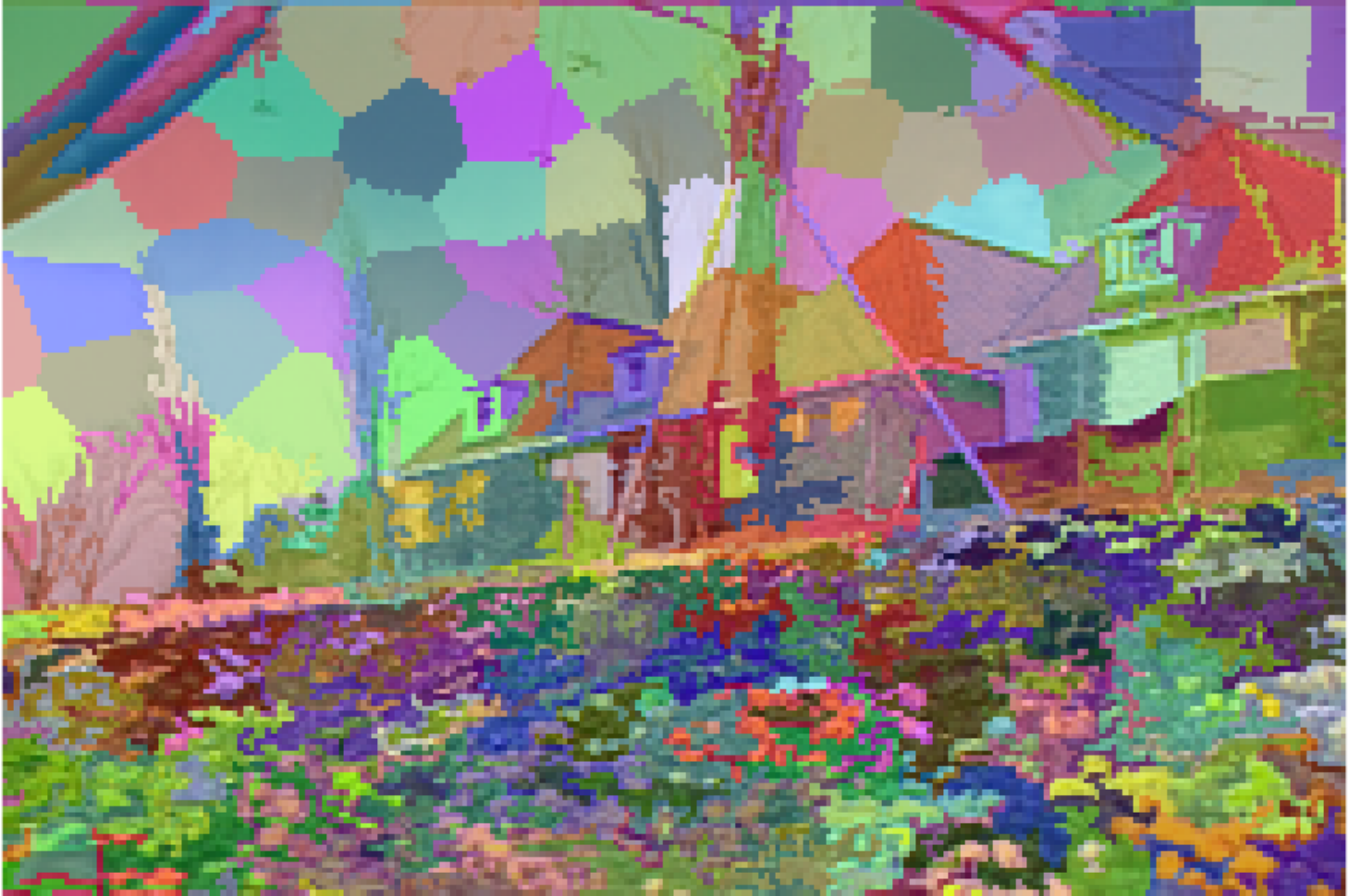}
\includegraphics[width=\w]{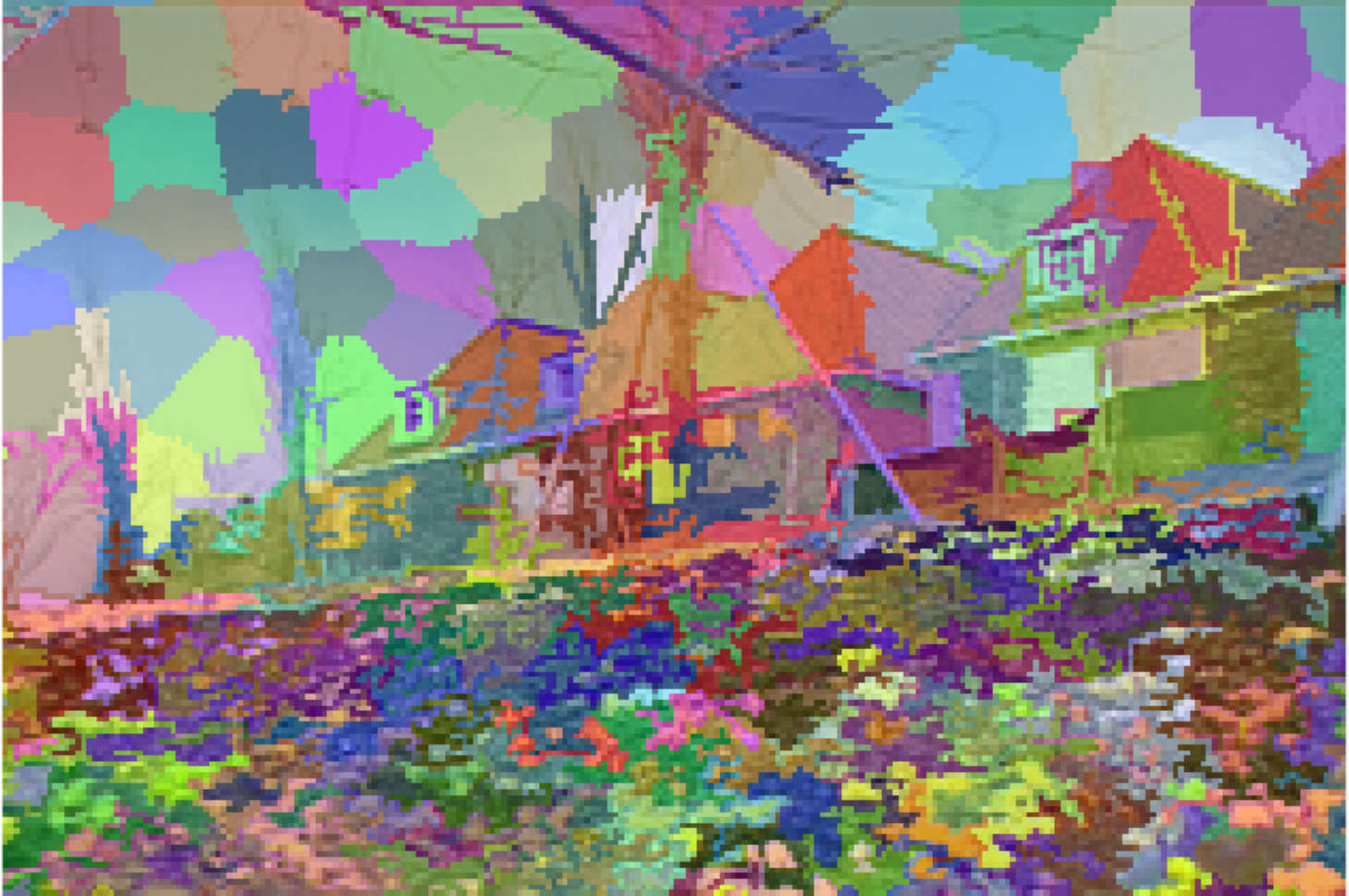}

\includegraphics[width=\w]{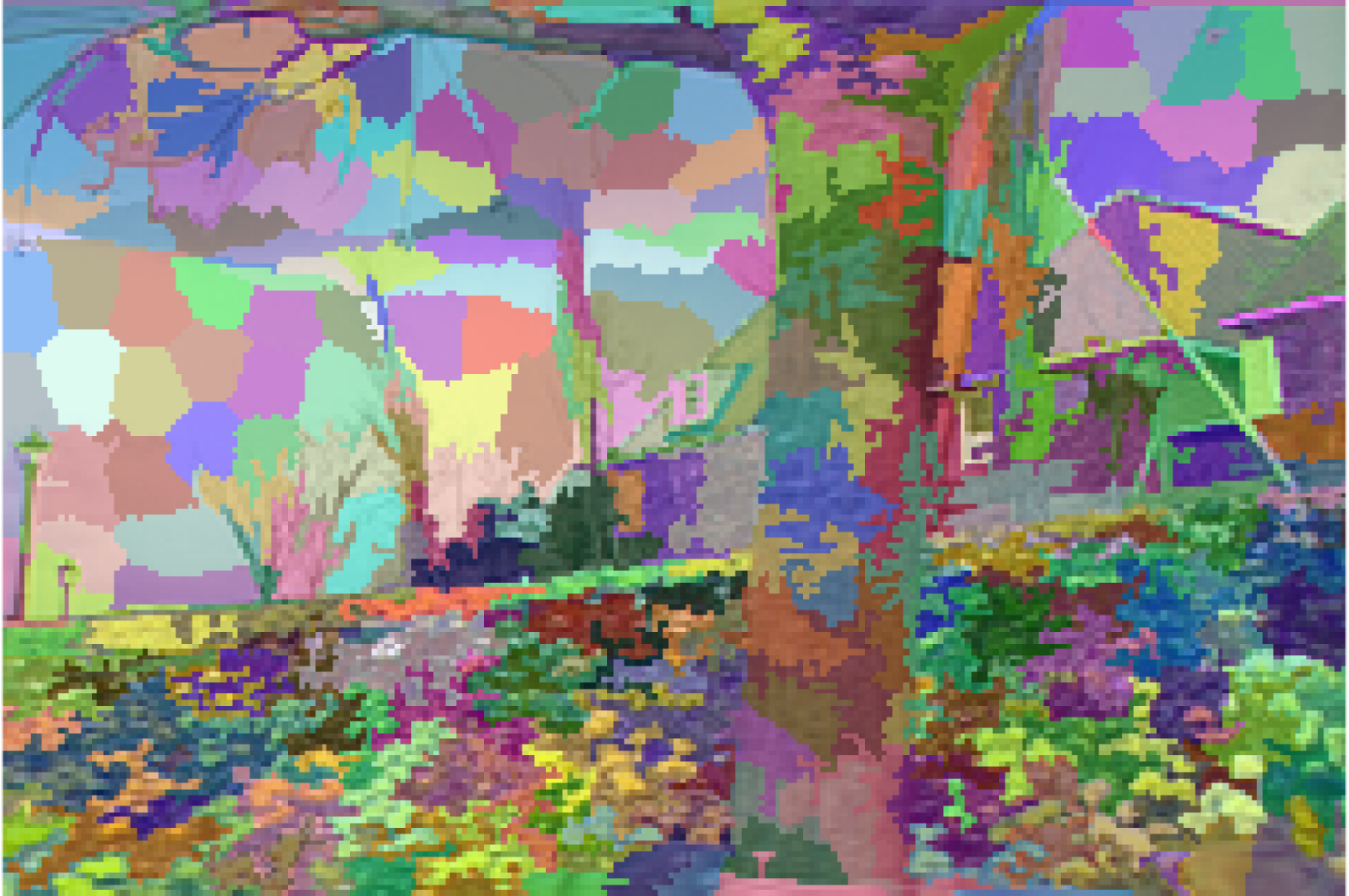}
\includegraphics[width=\w]{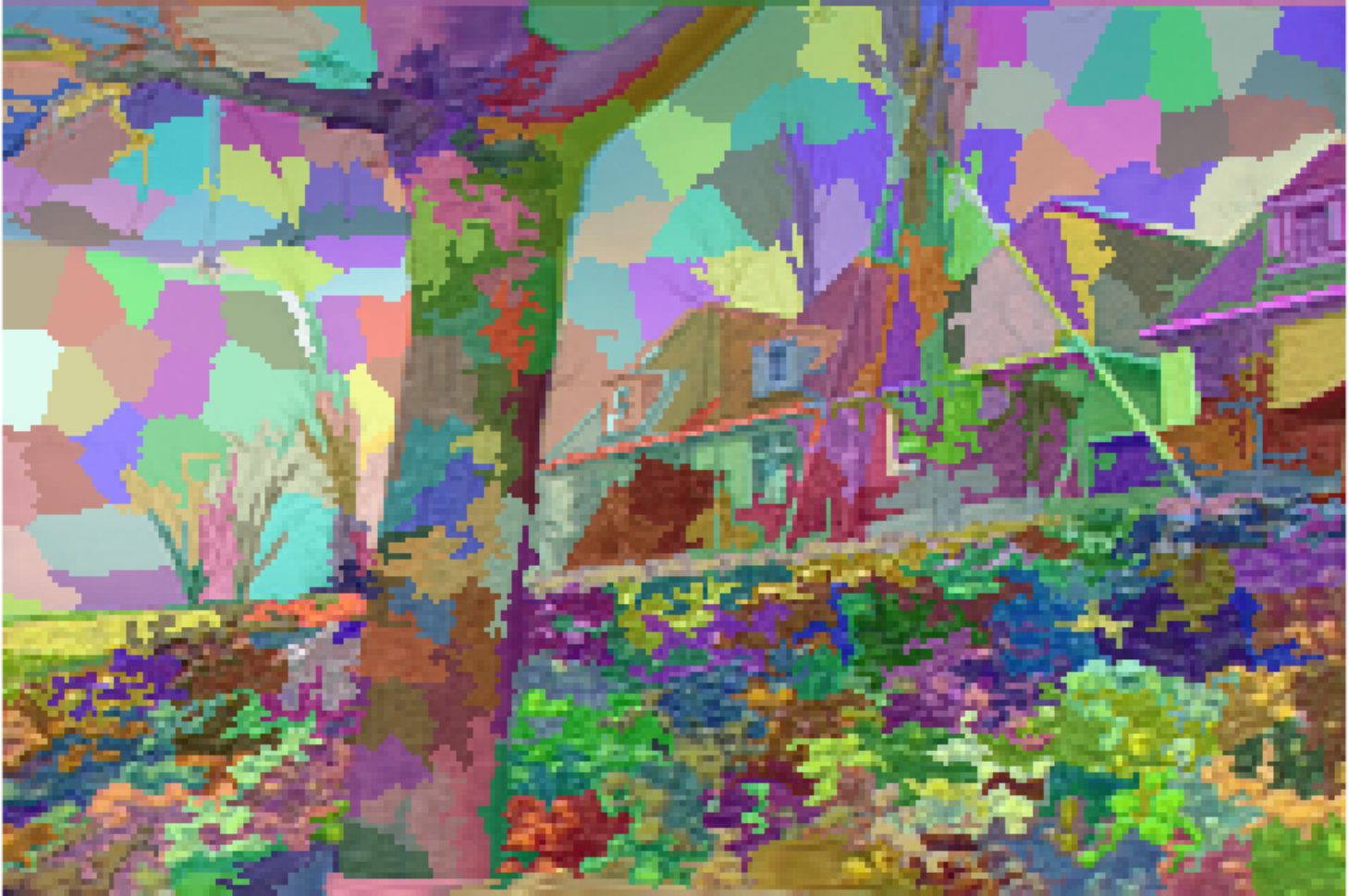}
\includegraphics[width=\w]{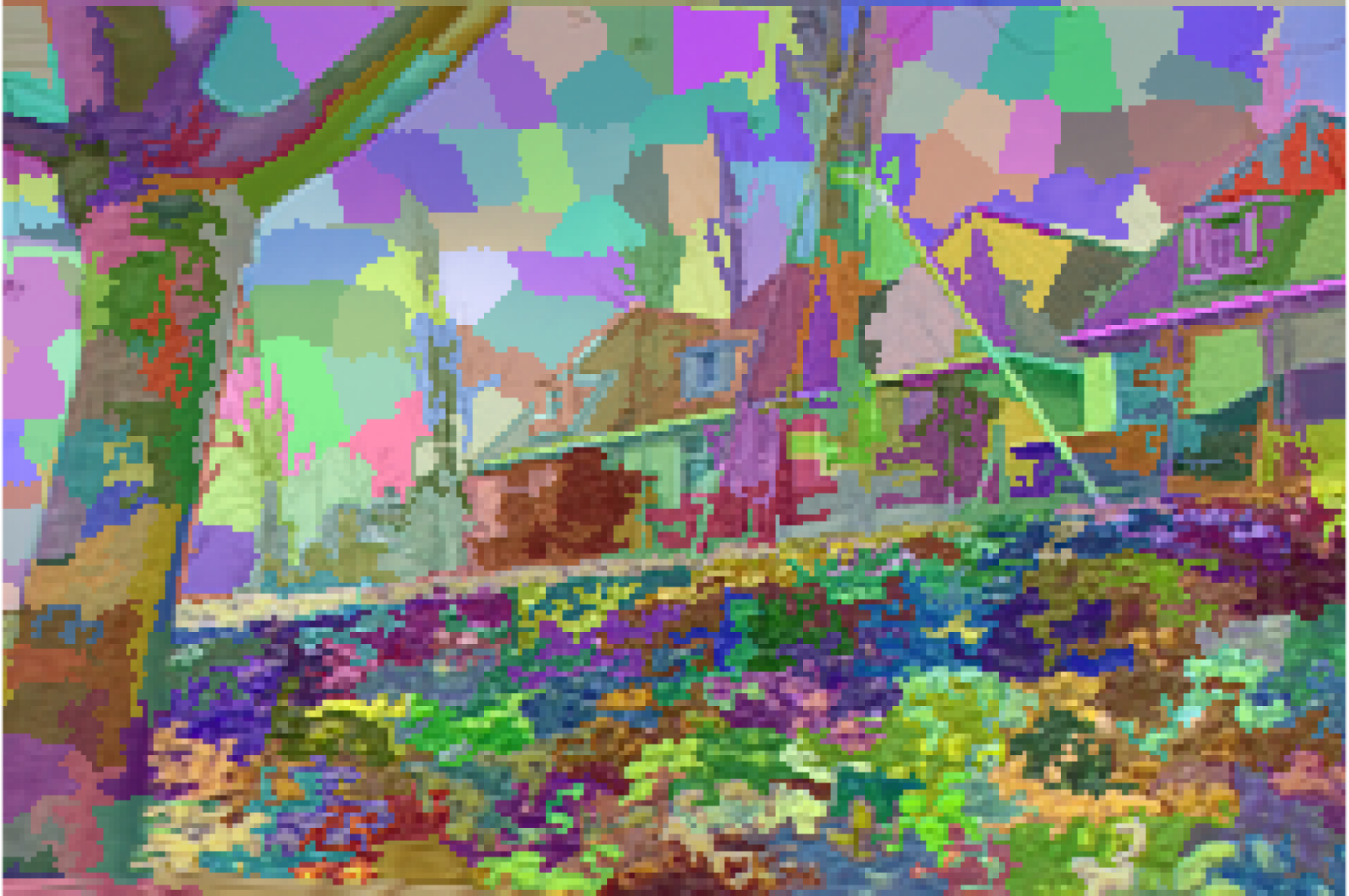}
\includegraphics[width=\w]{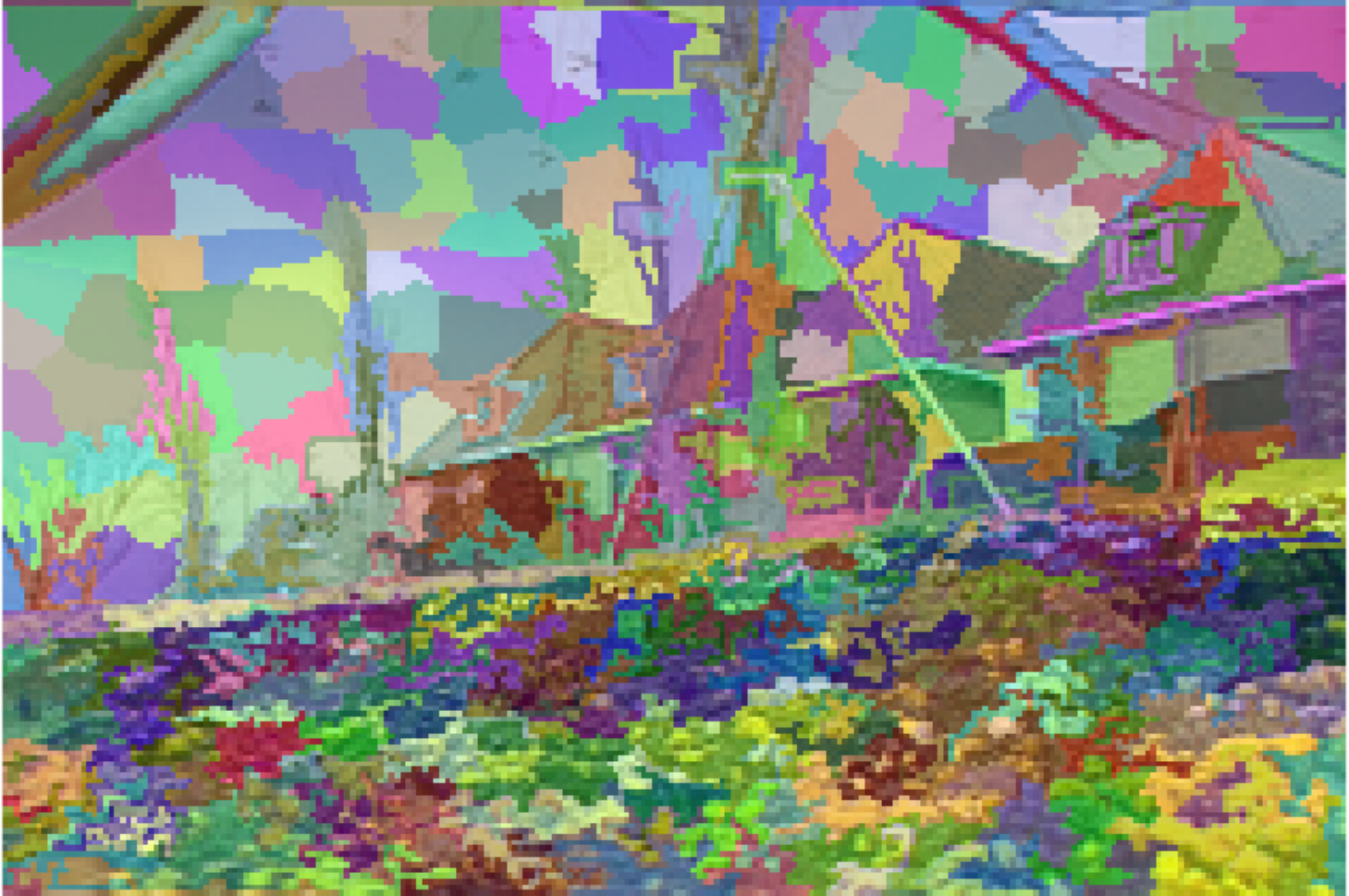}
\includegraphics[width=\w]{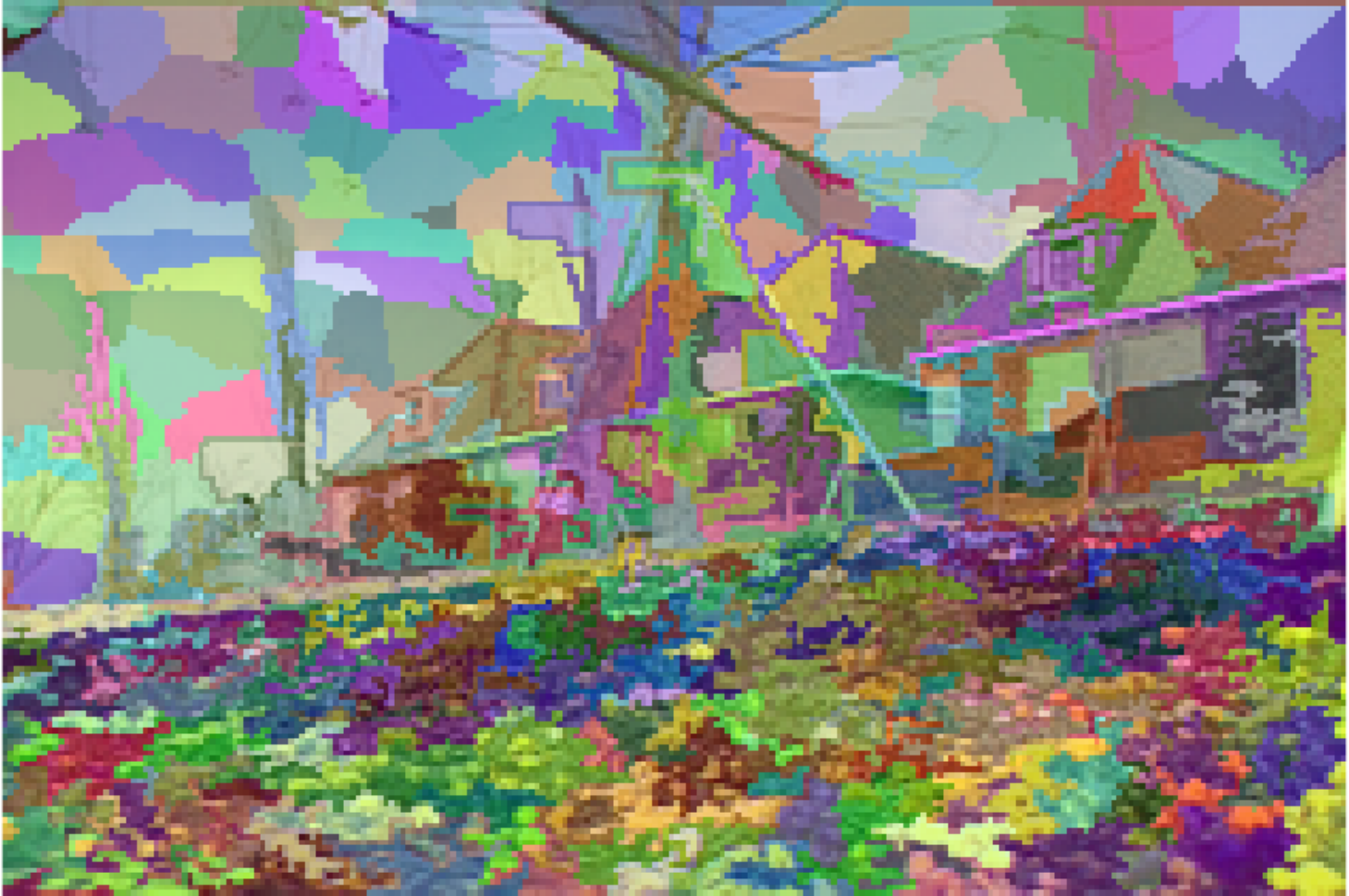}

\includegraphics[width=\w]{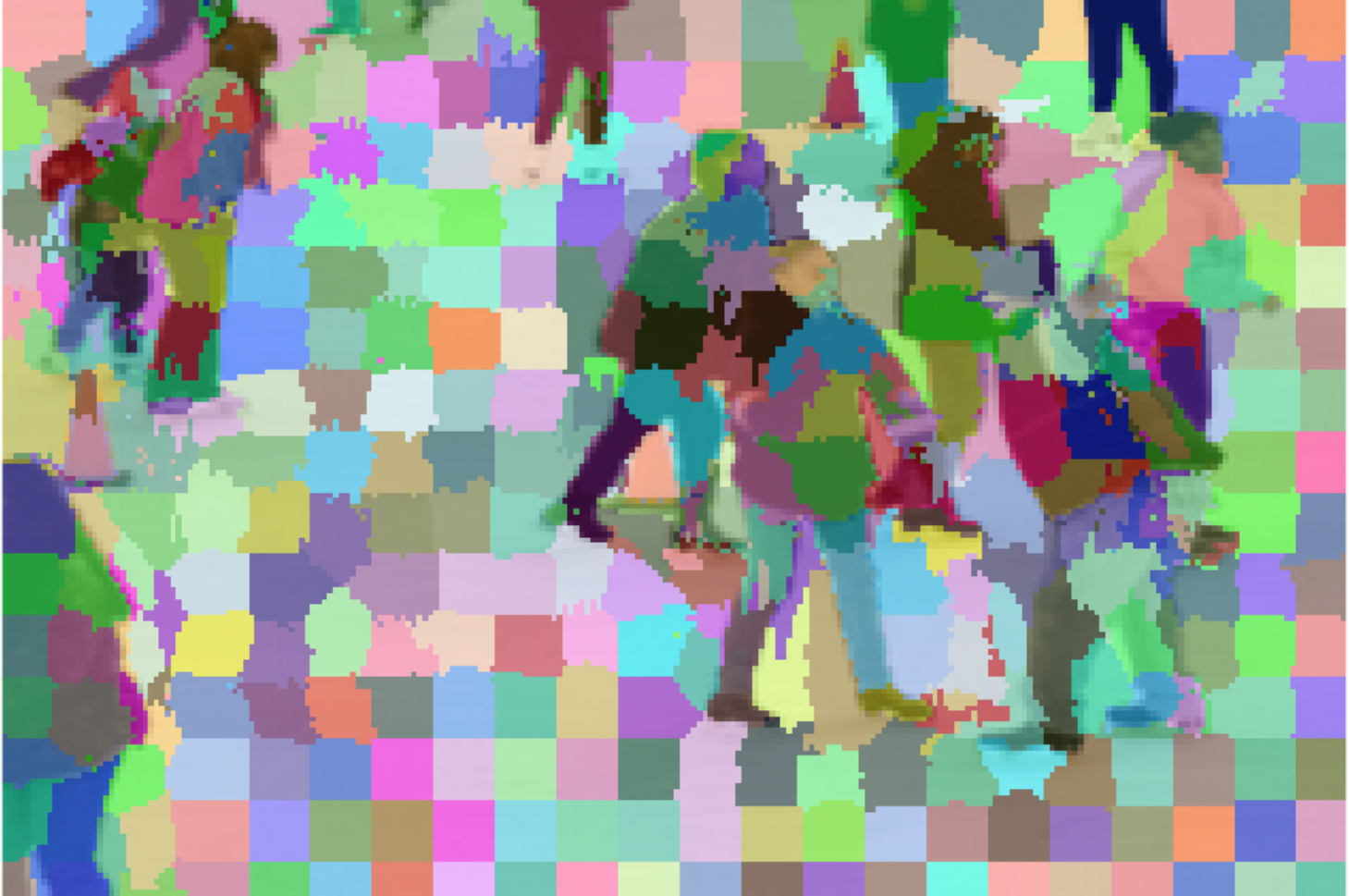}
\includegraphics[width=\w]{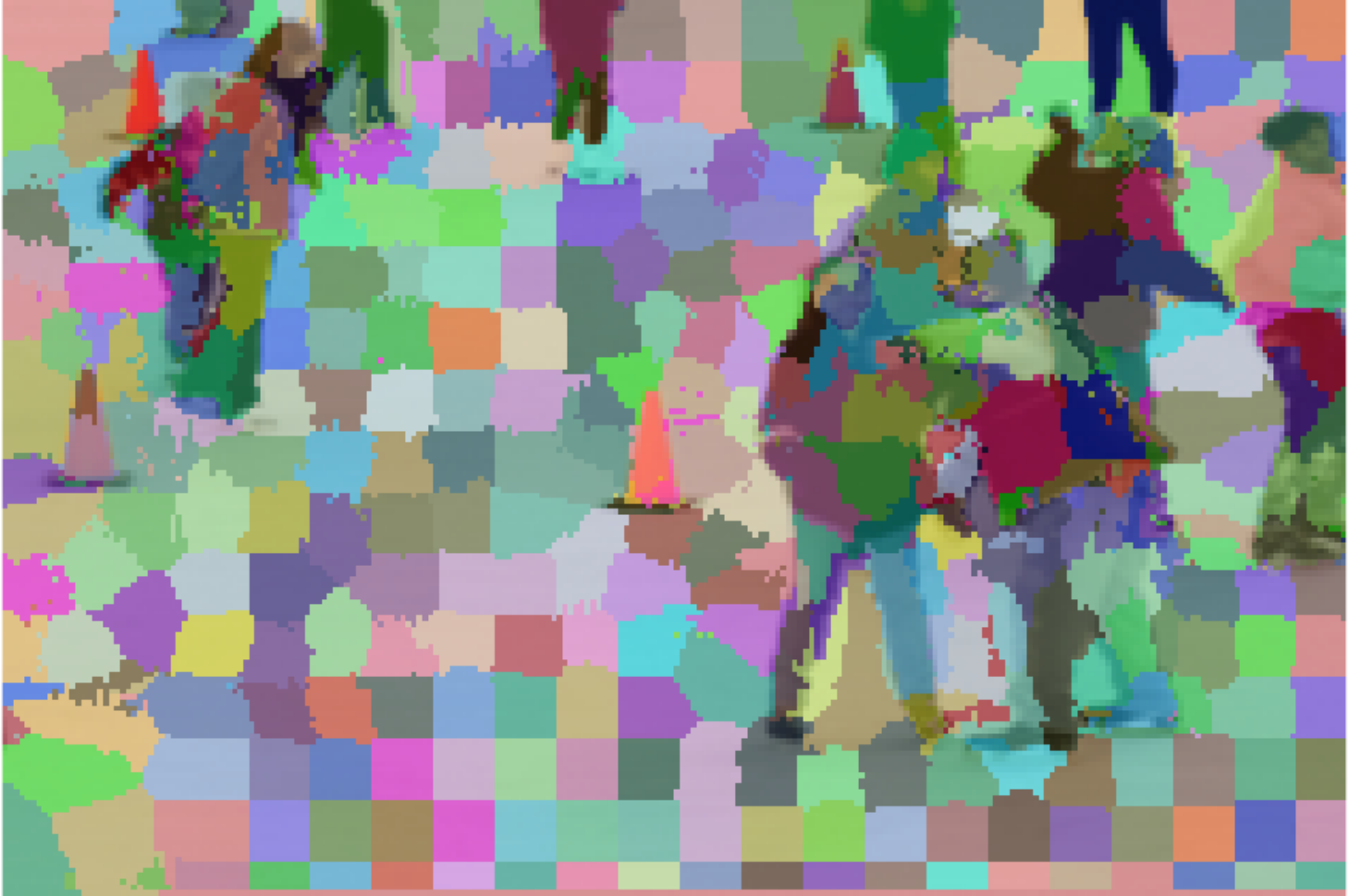}
\includegraphics[width=\w]{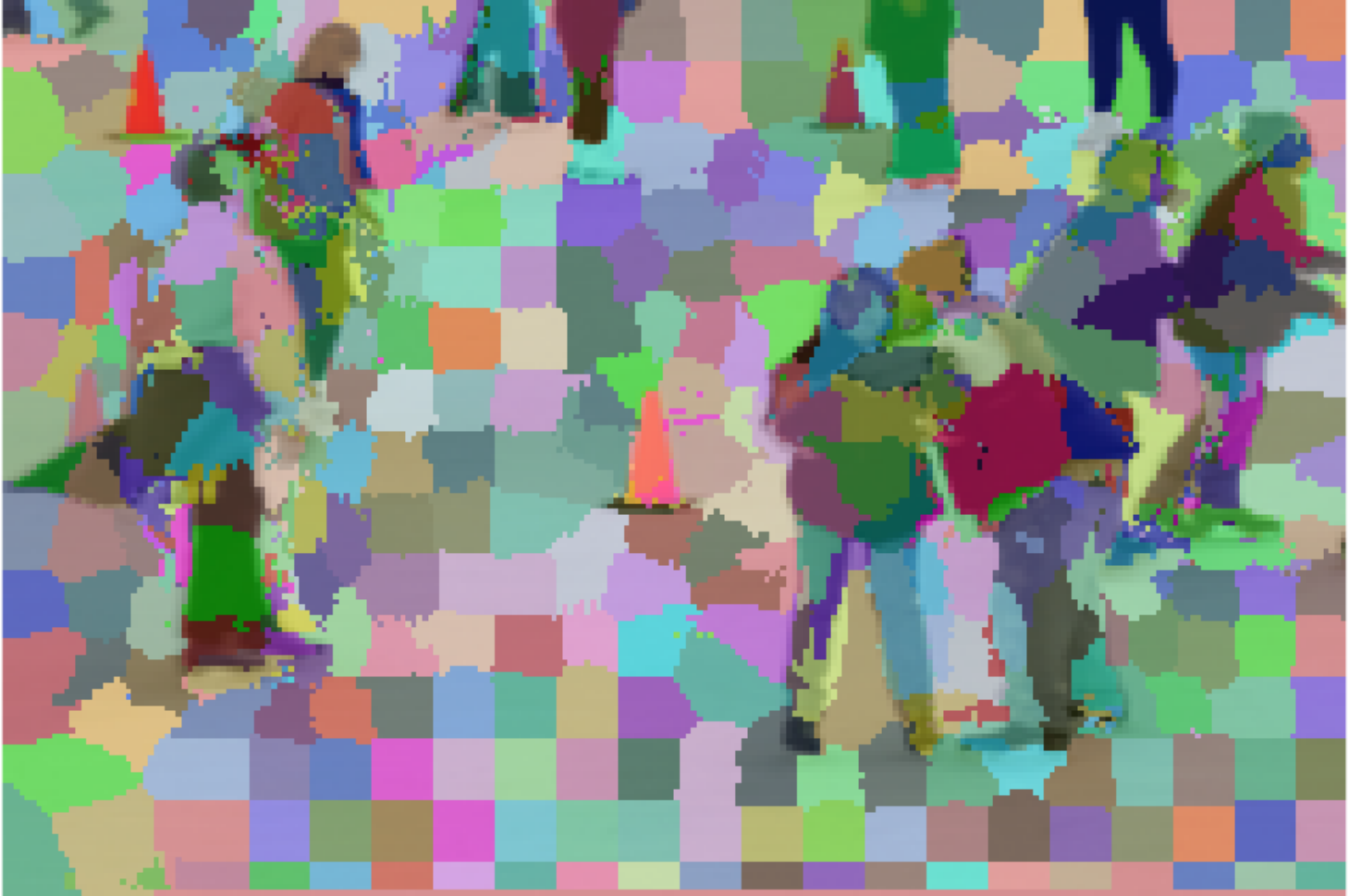}
\includegraphics[width=\w]{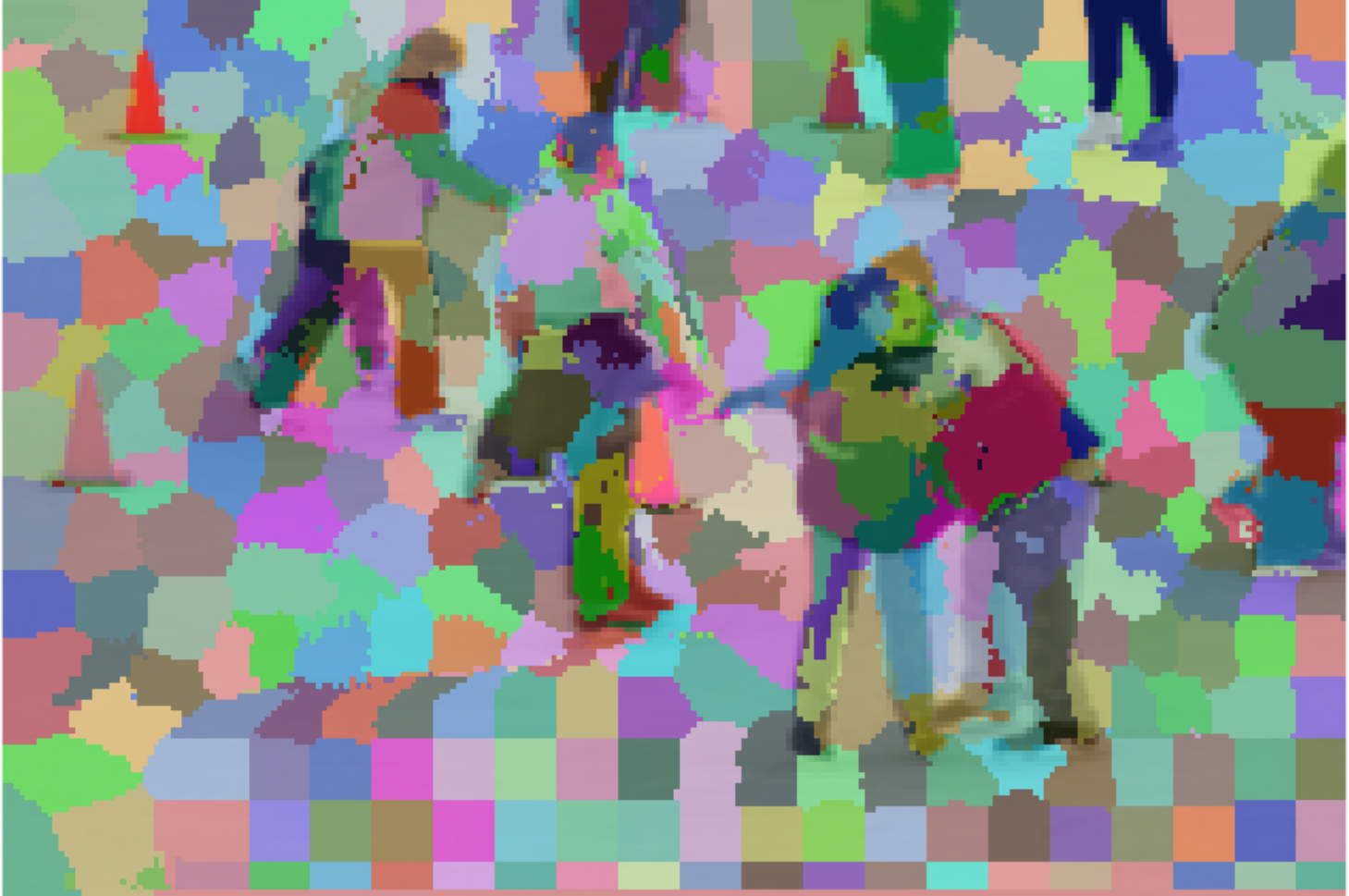}
\includegraphics[width=\w]{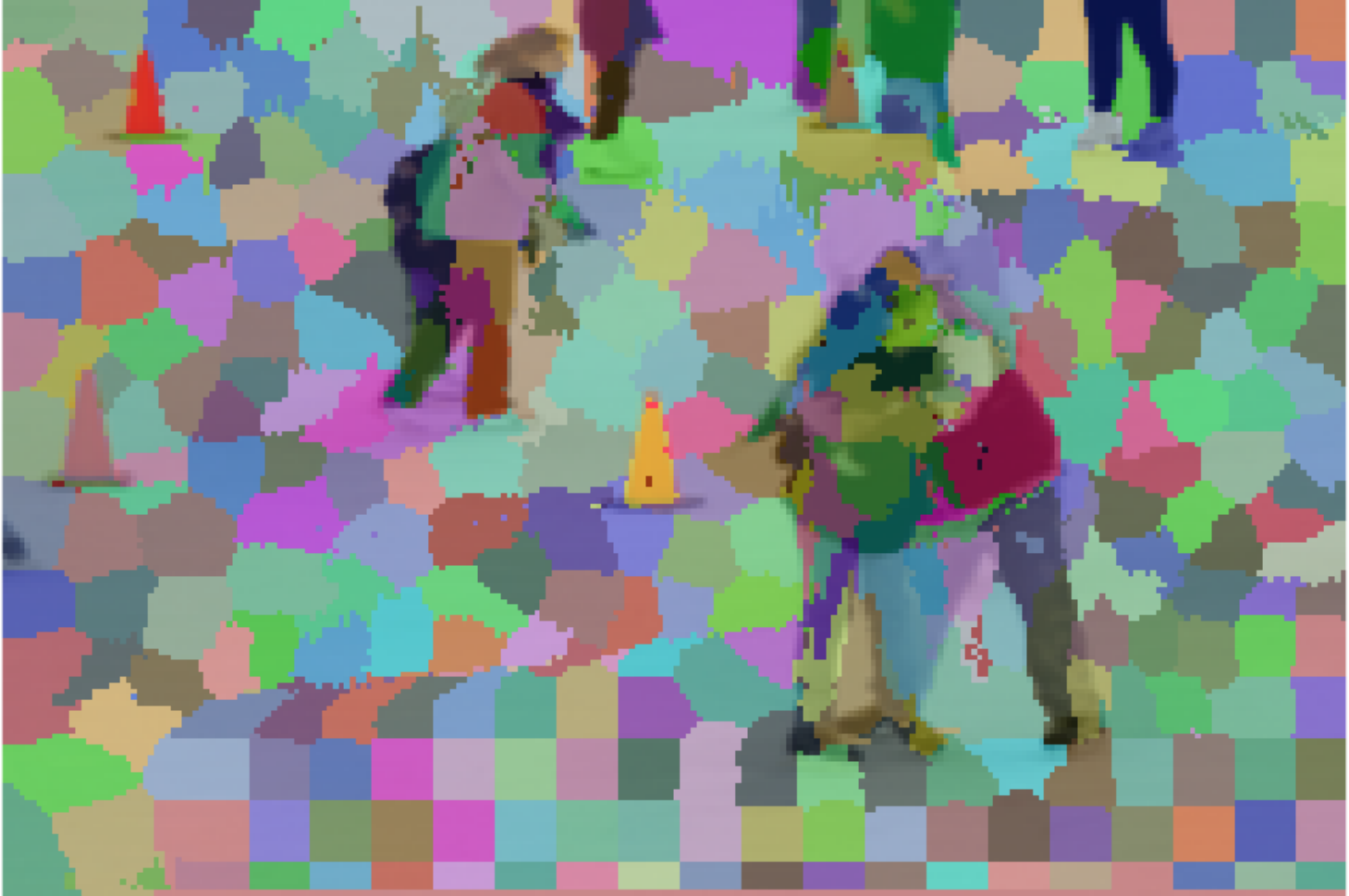}

\includegraphics[width=\w]{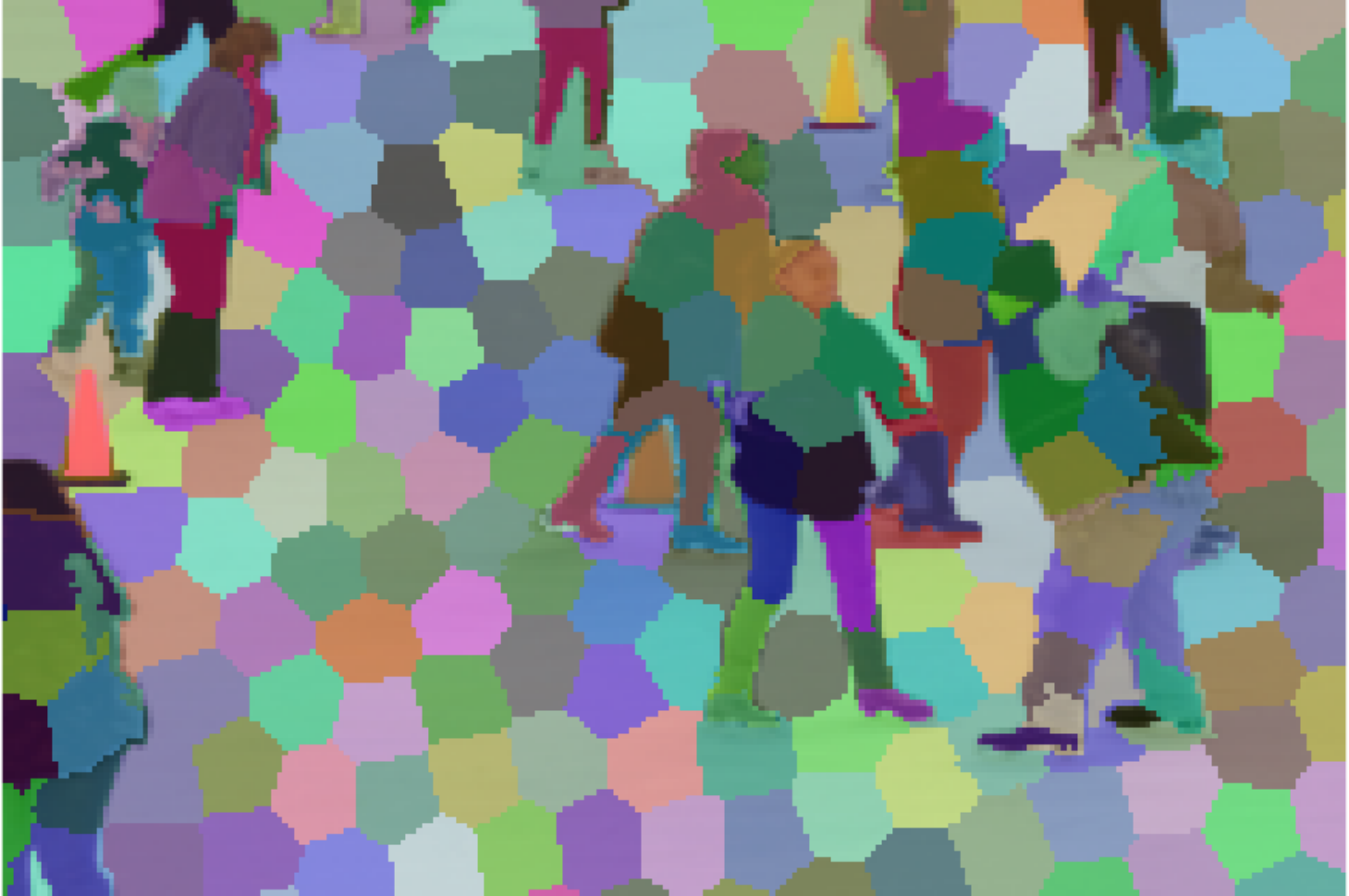}
\includegraphics[width=\w]{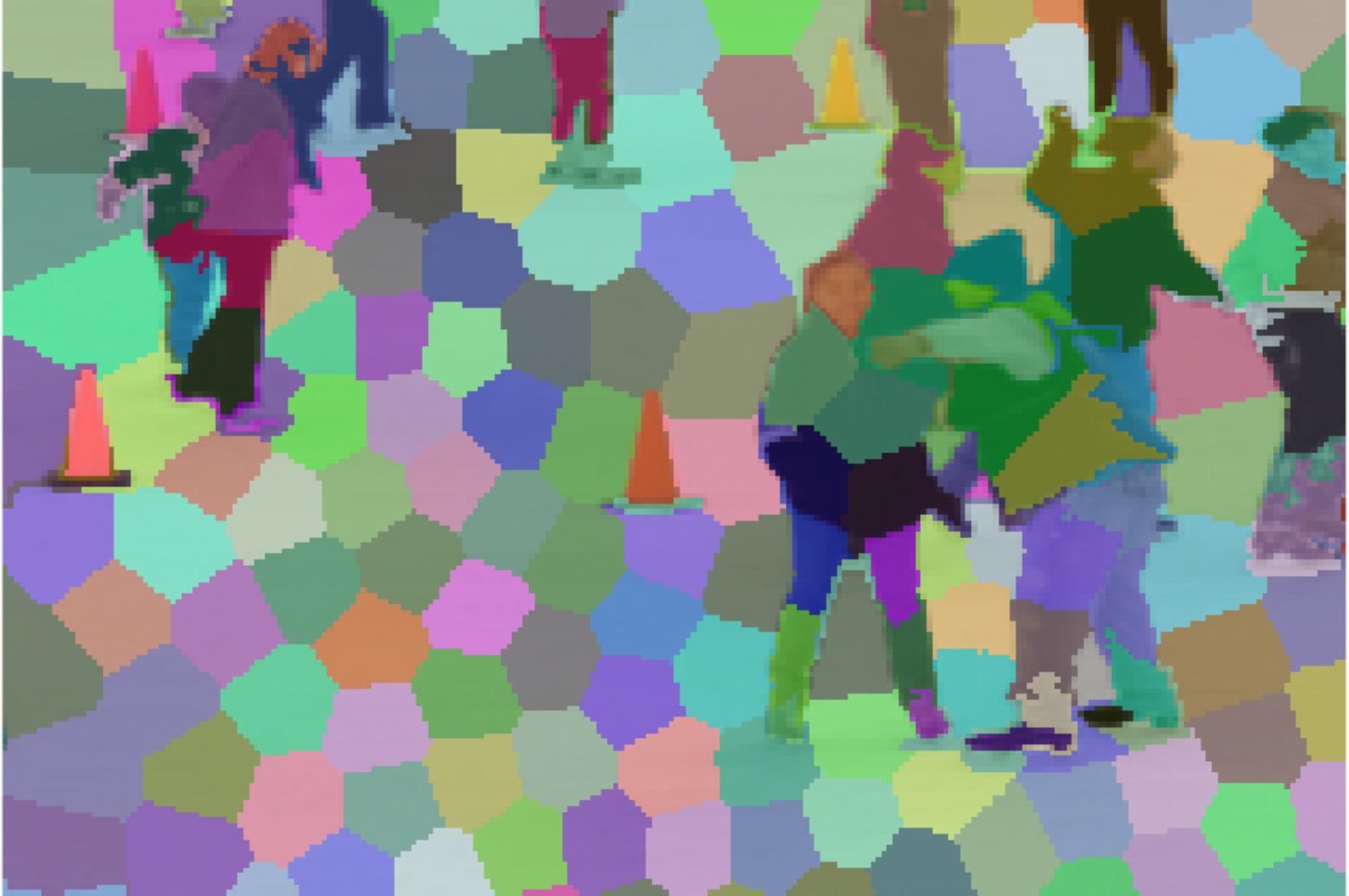}
\includegraphics[width=\w]{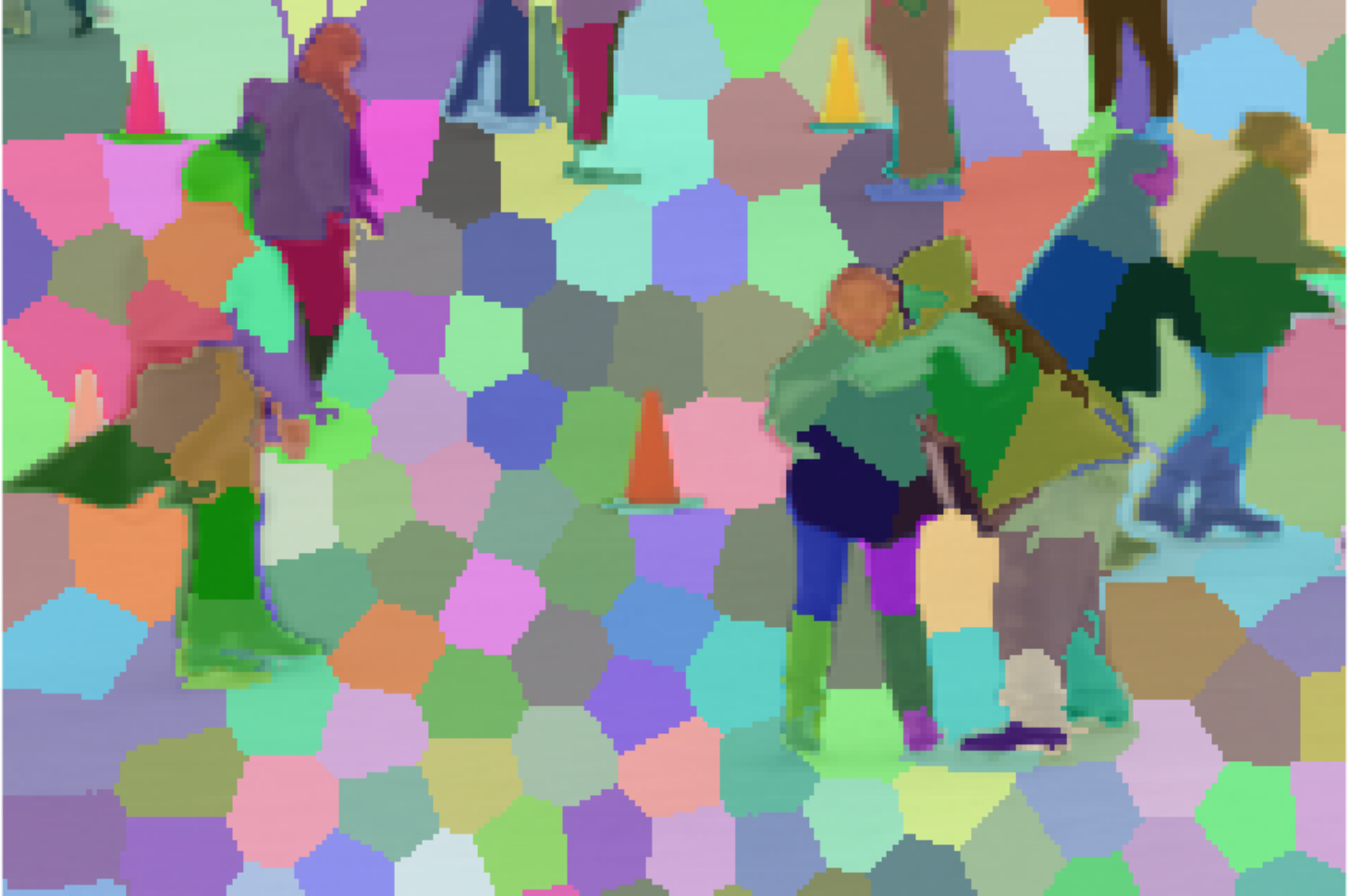}
\includegraphics[width=\w]{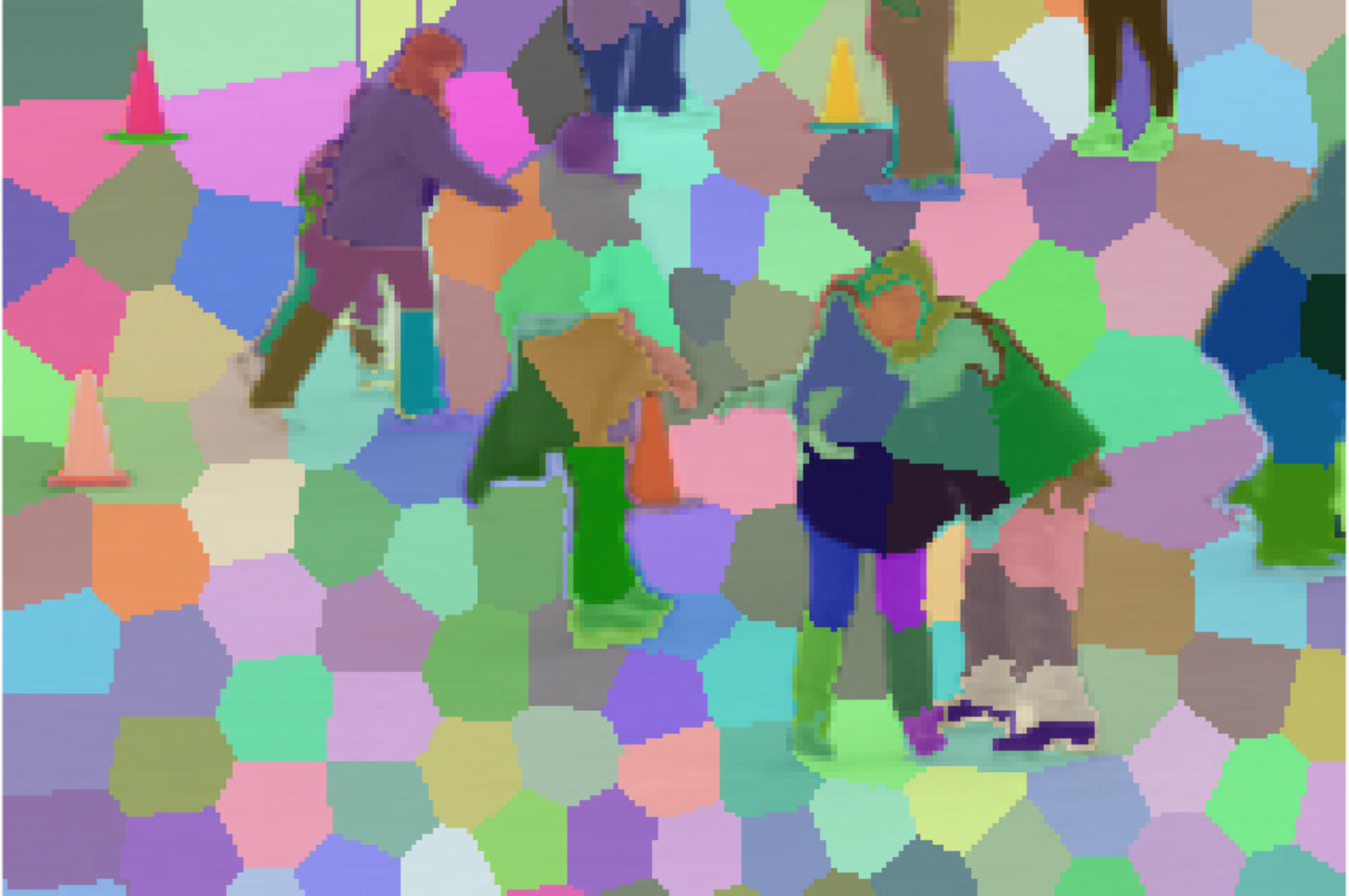}
\includegraphics[width=\w]{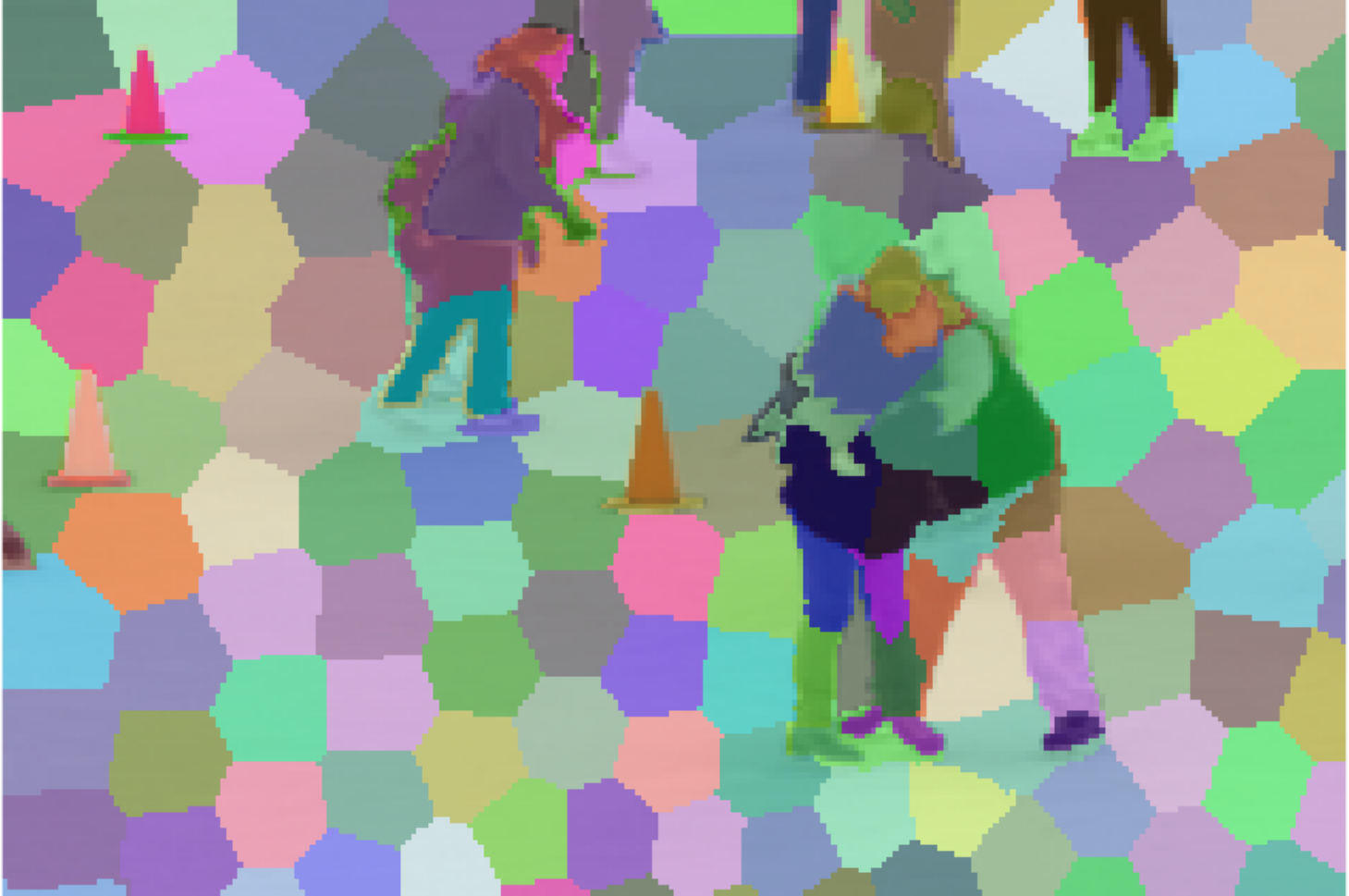}

\includegraphics[width=\w]{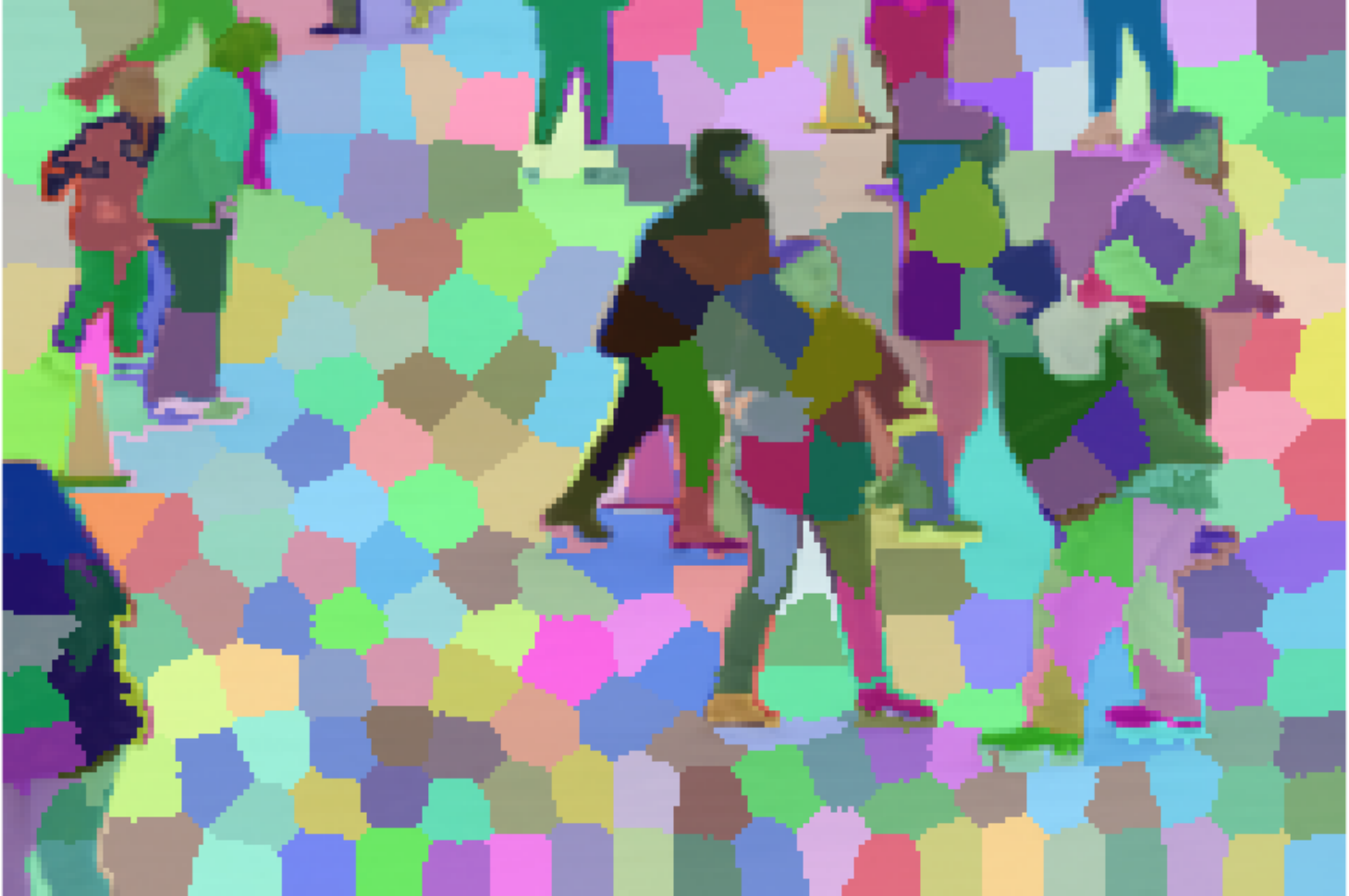}
\includegraphics[width=\w]{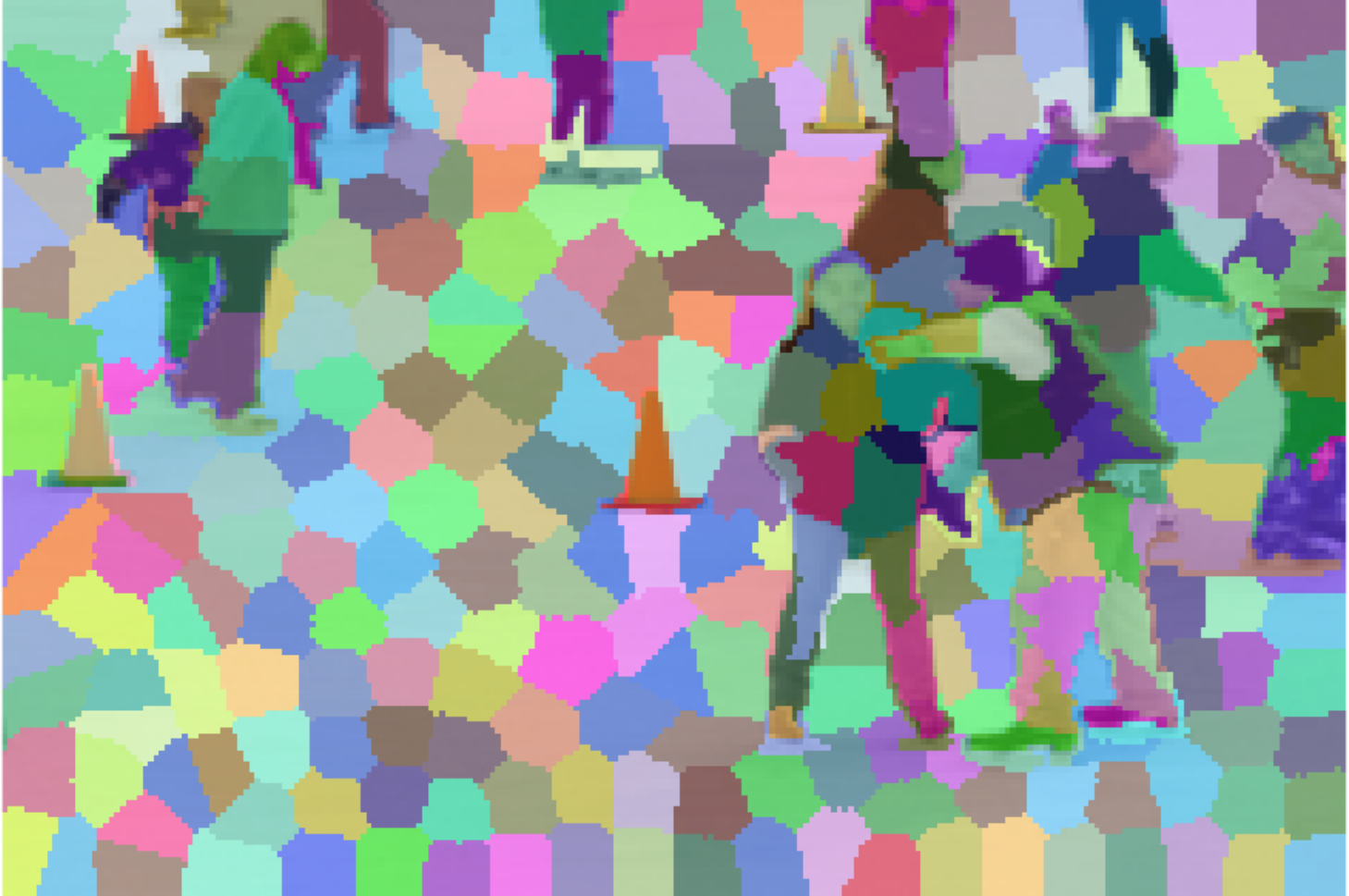}
\includegraphics[width=\w]{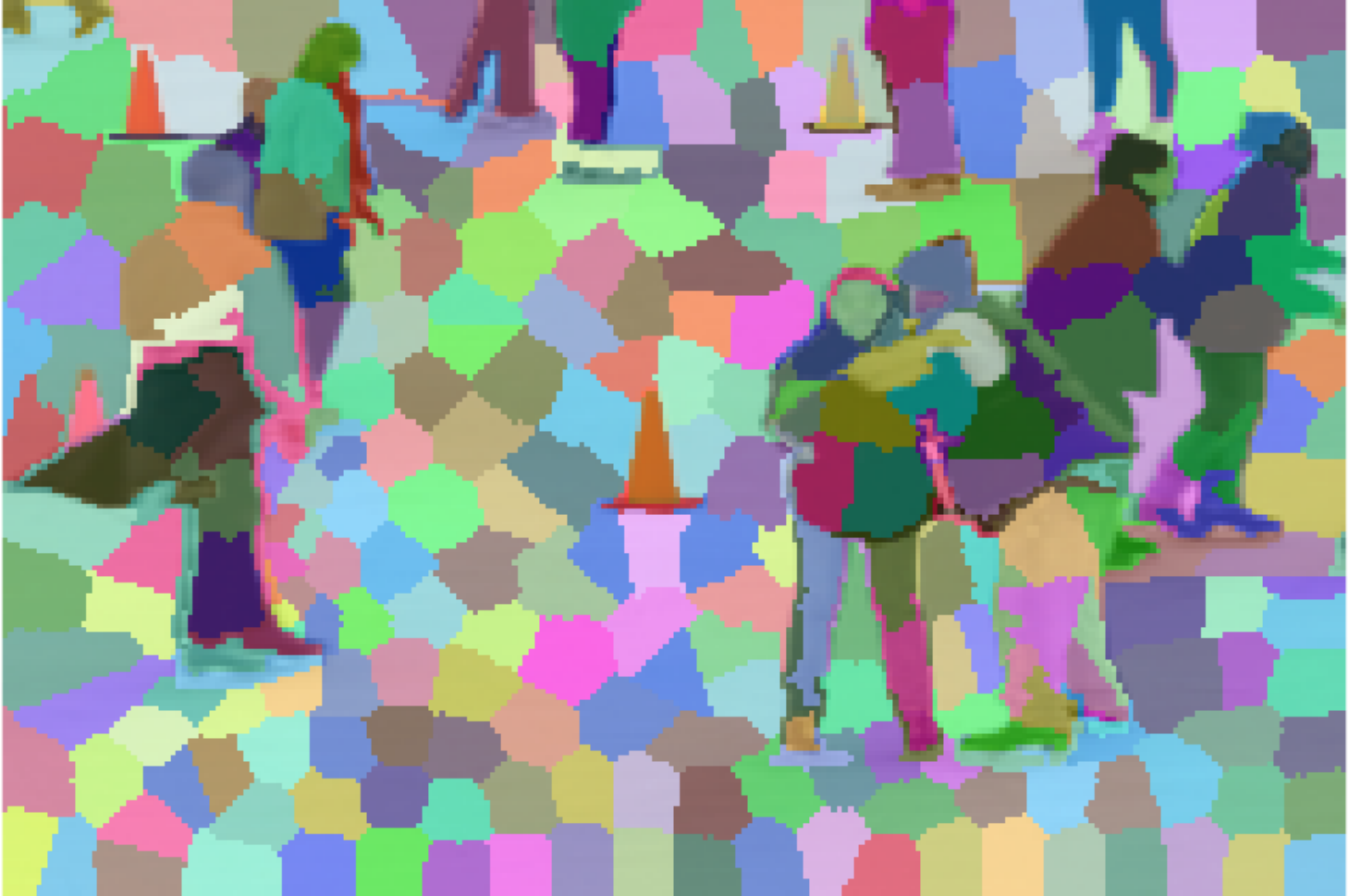}
\includegraphics[width=\w]{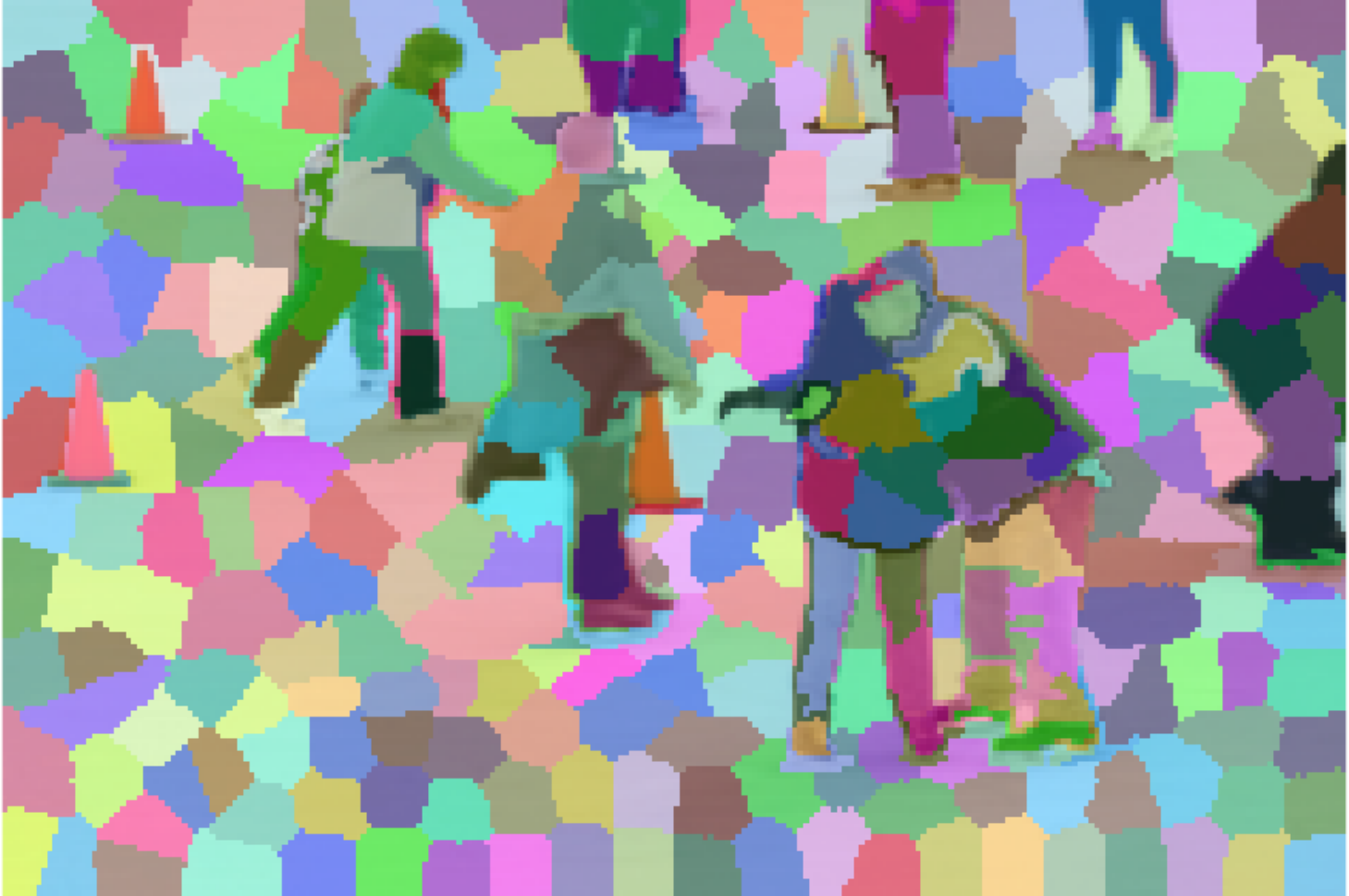}
\includegraphics[width=\w]{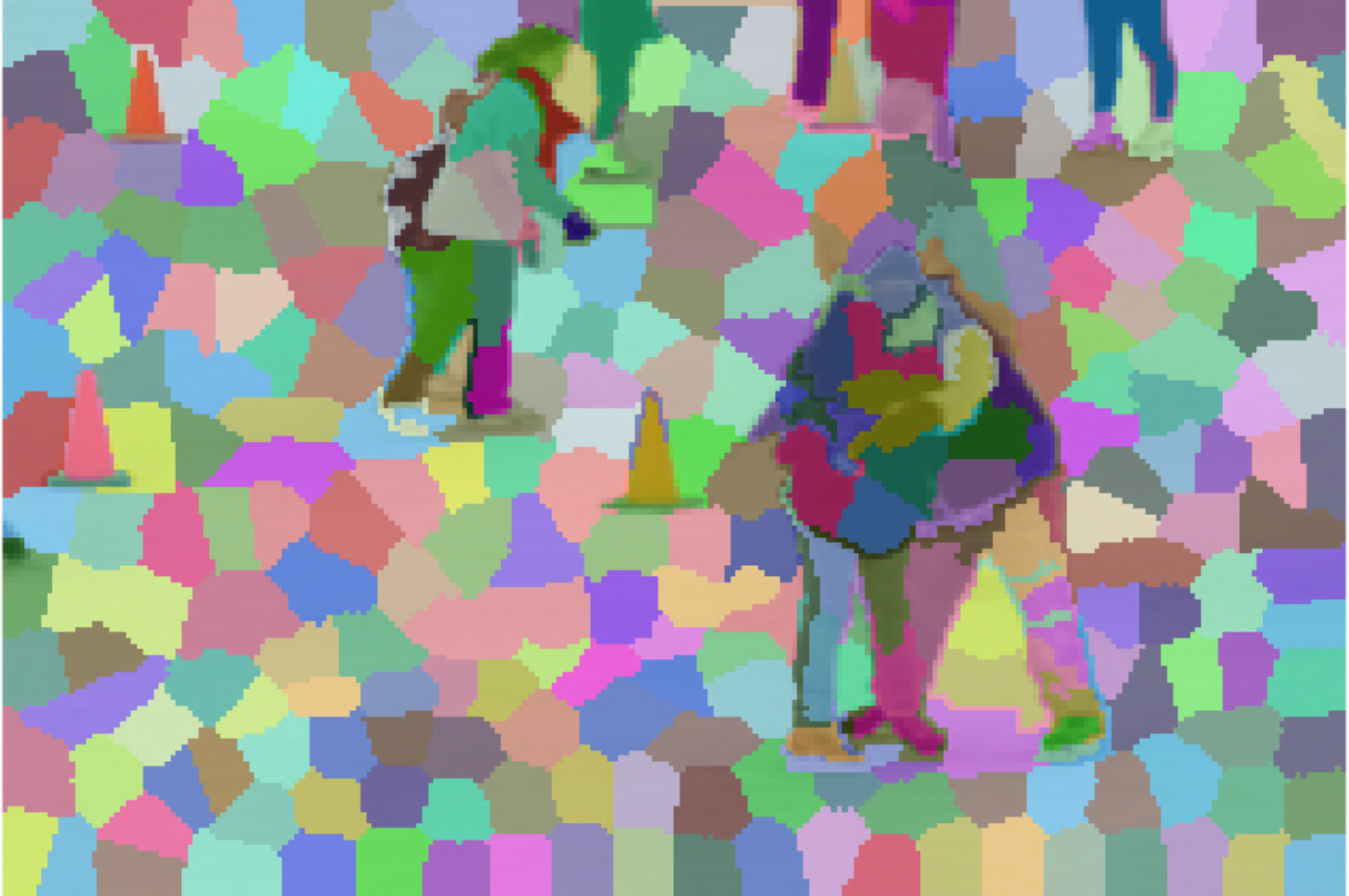}

   \caption{Qualitative evaluation of the proposed method ($1^\mathrm{st}$ and $4^\mathrm{th}$ rows), TSP \cite{chang2013video} ($2^\mathrm{nd}$ and $5^\mathrm{th}$ rows) and CCS \cite{lee2017contour} ($3^\mathrm{rd}$ and $6^\mathrm{th}$ row) on Garden and Ice sequence from the Chen dataset \cite{chen2010propagating}, over the span of 80 frames for roughly equal number of supervoxels. Different colors show different temporal superpixels. All three algorithms are able to track superpixels on both rigid and nonrigid objects over the whole video. (Best viewed in color)}
\label{fig:qualitative}
\end{figure}

%Although trajectory based clustering formulation provides global consistency, its inherent drawback is the accumulation of error in the trajectory tracking. Hence, unless the tracking error is small, the previous methods not using trajectories are expected to give better temporal superpixels than the proposed method. Nonetheless, the advantage of super-trajectories is their use in trajectory based video analysis like motion segmentation or activity forecasting. Whereas using temporal superpixels for such applications is not straight forward. 

As we discussed, our main contribution is a new video representation, which should be useful for applications involving trajectory based video analysis, but here we also compare the segmentation accuracy of super-trajectories (ST) against two state of the art temporal superpixels methods: turbo superpixels (TSP) \cite {chang2013video} and contour-constrained superpixels (CCS) \cite{lee2017contour}. We use Full-flow \cite{chen2016full} for optical flow estimation and convert this to trajectories using the method discussed in Section \ref{sec:OF2Traj} and then cluster them using Algorithm \ref{alg:STrj}. Given the trajectory labels, we estimate the pixel labels at a frame $f$ using Equation \ref{Eq:g}. The estimated pixel labels are compared against the temporal superpixels obtained from TSP and CCS. To ensure fairness, we also use the optical flow produced by Full-flow for TSP and CCS. We use LIBSVX 3.0 benchmark \cite{xu2012evaluation} to evaluate the accuracy of temporal superpixels. We use the seven evaluation metrics in \cite{xu2012evaluation}: 2D undersegmentation error (UE2D), 2D segmentation accuracy (SA2D), 2D boundary recall (BR2D), 3D undersegmentation error (UE3D), 3D segmentation accuracy (SA3D), 3D boundary recall (BR3D), and the mean duration against the desired average superpixel size, $s$. In addition we also count the number of supervoxels for each algorithm.

Fig \ref{fig:quantitative} compares the quantitative results on the Chen dataset \cite{chen2010propagating}, against the number of supervoxels for each algorithm. This dataset consists of eight videos, each of length roughly equal to 80 frames and 5-10 ground-truth objects per video. The figure shows that the proposed method gives better UE2D and UE3D than the previous methods. In addition the mean duration of our method is longer and it gives fewer number of supervoxels than the previous methods but other evaluation metrics, we are slightly worse off. To further investigate the source of error, in Fig \ref{fig:qualitative}, we give a qualitative comparison of all three methods on a few frames on Garden and Ice sequence from Chen dataset. All three methods give accurate boundaries for most of the superpixels. and the places where ST did mistakes are very few (See red arrows in the figure). The source of these mistakes is error in optical flow. Most optical flow methods exhibit a shrinking bias: they tend to be inaccurate toward small or elongate objects (such as the skater's legs or thin tree branches). The reason CCS and TSP perform better is because they track at superpixels level whereas we maintain pixel level tracking in the form of trajectories and suffer from the accumulation of error. However the proposed method is able to track superpixel for longer durations. For a better qualitative comparison, please refer to the supplementary videos.  In addition, the main benefit of our approach is that super-trajectories contain more information than temporal superpixels in the form of dense pixel-level tracking. With more accurate trajectories, results could be improved.

%Fig \ref{fig:qualitative} shows our super-trajectories and the temporal superpixels of TSP and CCS for roughly equal number of supervoxels for both the algorithms. All three algorithms are able to track superpixels for both rigid and nonrigid objects over the whole video.

%=====================================================================================
\section{Conclusion} We propose super-trajectories, an over-segmentation of dense trajectories as a new representation for videos. The representation in terms of trajectories rather than pixels imposes long temporal consistency in a global manner. The proposed algorithms can be used to find a clustering of trajectories having minimum dis-connections among them. Super-trajectories have applications in trajectory based video analysis. We hope that the Vision community would find the proposed representation useful in a number of applications.

%\begin{figure}
%  \centering
%  \fbox{\rule[-.5cm]{4cm}{4cm} \rule[-.5cm]{4cm}{0cm}}
%  \caption{Sample figure caption.}
%  \label{fig:fig1}
%\end{figure}

%\begin{table}
% \caption{Sample table title}
%  \centering
%  \begin{tabular}{lll}
%    \toprule
%    \multicolumn{2}{c}{Part}                   \\
%    \cmidrule(r){1-2}
%    Name     & Description     & Size ($\mu$m) \\
%    \midrule
%    Dendrite & Input terminal  & $\sim$100     \\
%    Axon     & Output terminal & $\sim$10      \\
%    Soma     & Cell body       & up to $10^6$  \\
%    \bottomrule
%  \end{tabular}
%  \label{tab:table}
%\end{table}
%
%\subsection{Lists}
%\begin{itemize}
%\item Lorem ipsum dolor sit amet
%\item consectetur adipiscing elit. 
%\item Aliquam dignissim blandit est, in dictum tortor gravida eget. In ac rutrum magna.
%\end{itemize}

\bibliographystyle{unsrt}  
\bibliography{egbib}  %%% Remove comment to use the external .bib file (using bibtex).
%%% and comment out the ``thebibliography'' section.

%%% Comment out this section when you \bibliography{references} is enabled.
%\begin{thebibliography}{1}
%\bibliography{egbib}
%\end{thebibliography}

\end{document}